\documentclass[lettersize,journal]{IEEEtran}
\IEEEoverridecommandlockouts
\usepackage{cite}
\usepackage{amsmath,amssymb,amsfonts}
\usepackage{algorithmic}
\usepackage{graphicx}
\usepackage{textcomp}
\usepackage{bm}
\usepackage{multirow}
\usepackage{supertabular}
\usepackage[T1]{fontenc}
\usepackage{longtable}
\usepackage{xcolor}
\usepackage{subcaption}
\usepackage{tabularx}
\usepackage{comment}
\usepackage{booktabs}
\usepackage{mathrsfs}
\usepackage{float}
\usepackage{cuted}
\usepackage{algorithm}

\allowdisplaybreaks
\def\BibTeX{{\rm B\kern-.05em{\sc i\kern-.025em b}\kern-.08em
    T\kern-.1667em\lower.7ex\hbox{E}\kern-.125emX}}

\begin{document}

\title{Human-Feedback-Guided Multi-Agent Inverse Reinforcement Learning for Interactive Driving Behavior Modeling}

\author{Longhao Yan, Jinhao Liang, Kaidi Yang
\thanks{The authors are with the Department of Civil and Environmental Engineering, National University of Singapore, Singapore 117576. Email:{longhao.yan@u.nus.edu, jh.liang0523@gmail.com, kaidi.yang@nus.edu.sg}. \textit{Corresponding author}: Kaidi Yang. }
\thanks{This research was supported by the Singapore Ministry of Education (MOE) under its AcRF Tier 1 Grant (A-8000404-01-00). 
}}

\maketitle

\begin{abstract}
Modeling human driving behavior is fundamental to understanding driver decision-making and supporting autonomous vehicles (AVs) in mixed traffic. Reward learning offers a principled way to infer the latent objectives behind driving behaviors, yet most existing methods are formulated in single-agent settings and treat surrounding vehicles as dynamic obstacles. Directly extending reward learning to multi-agent scenarios is difficult because the joint action space grows exponentially with the number of interacting vehicles. In addition, existing approaches mainly rely on physical demonstrations and rarely exploit human feedback during reward recovery. To address these limitations, this paper proposes a Human-Feedback-Guided Multi-Agent Inverse Reinforcement Learning (HFG-MAIRL) framework for interactive driving behavior modeling. First, we develop a multi-agent maximum-entropy inverse reinforcement learning module that explicitly models each vehicle and its directly interacting neighbors in a localized Markov game, while using aggregated traffic states to capture the influence of the broader traffic environment. The resulting coupled multi-vehicle responses are approximated through finite marginal-response iterations over factorized vehicle-level policies, avoiding explicit enumeration of the full joint action space. Second, we incorporate human feedback as ordinal preference supervision through a Bradley–Terry likelihood. In our experiments, driver emotion labels are used as one instantiation of such feedback and converted into pairwise ordinal preference constraints. Experiments on the AIDE dataset show that HFG-MAIRL outperforms representative IRL baselines in decision modeling and produces interpretable feedback-aligned reward patterns.
\end{abstract}


\section{Introduction}

\IEEEPARstart{A}{lthough} autonomous vehicles (AVs) promise to transform future mobility, their large-scale deployment remains challenged by the inherently human-oriented nature of transportation systems. 
In such systems, AVs operate alongside human-driven vehicles (HDVs) in mixed traffic, with their decision-making shaped by interactions with surrounding human drivers and the social expectations embedded in traffic environments. This calls for a principled understanding of human driving behavior at two complementary levels~\cite{wang2023adaptive,garg2023can,hang2023brain,wang2023deep,lu2022instance,lu2019transfer,zhang2024noncooperative,rasch2024safe}. First, understanding the behaviors of surrounding human drivers is essential for anticipating potential conflicts and enabling safer decision-making. Second, understanding the rationale underlying these behaviors is important for developing AV decision-making frameworks that are more human-like and socially compatible.

Most existing approaches to human driving behavior modeling focus on predicting the short-term intent of surrounding agents (e.g., drivers or human traffic participants) based on historical trajectories~\cite{bhattacharyya2018multi,sun2024modelling,liu2022interactive,zhang2025interactive,ye2025tp,zhao2025pmi,huang2026physics,yan2025bi,taghavifar2025behaviorally}. 
For instance, Ref.~\cite{liu2022interactive} proposed a deep learning framework integrated with a driving risk map that explicitly incorporates spatial risk priors to enhance interactive trajectory prediction of surrounding vehicles in highway scenarios. 
Ref.~\cite{zhang2025interactive} developed an encoder-decoder network architecture combined with parameterized transfer learning to capture complex vehicle interactions, improving the adaptability of the trajectory prediction model across different traffic domains. 
Ref.~\cite{ye2025tp} introduced a prediction framework that fuses global and local history information, tailored for complex unsignalized roundabouts, to capture multi-scale spatial-temporal interactions. 
However, these approaches do not explicitly capture the underlying decision-making process and may therefore generalize poorly to scenarios beyond their training data, such as behavioral shifts in response to AVs.

To uncover the underlying decision-making process behind driving behaviors, recent studies have attempted to infer, from expert demonstrations, the objectives or rewards that human drivers seek to optimize~\cite{sun2021adversarial,geng2023multimodal,kathuria2025learning,sun2022inverse,yang2024meta,wen2023modeling,schultner2025feature}. 
For example, Ref.~\cite{geng2023multimodal} applied inverse reinforcement learning (IRL) to infer risk-averse reward functions for human drivers at unsignalized intersections, explicitly modeling their trade-offs between safety and efficiency. 
Ref.~\cite{kathuria2025learning} utilized deep IRL to extract implicit reward maps from expert demonstrations, enabling agents to learn nuanced social navigation rules based on scene geometry.
Ref.~\cite{wen2023modeling} employed IRL to model human car-following dynamics, specifically quantifying how drivers adapt their underlying reward preferences when influenced by AVs. 
However, these approaches often consider single-agent settings, where each human driver is modeled in isolation, and surrounding agents are treated as part of the environment rather than strategic participants whose decisions mutually influence one another. 
As a result, the learned reward functions may not fully capture the strategic interactions that shape driving decisions in real-world traffic. 
Only a few recent studies have explicitly modeled multi-driver interactions through multi-agent IRL (MAIRL)~\cite{radtke2023predicting,zheng2021objective,liu2024sociality, li2025responsibility,lanzaro2025evaluating,mehrabi2026turbo}. 
For example, Ref.~\cite{zheng2021objective} framed driver interactions as a decentralized multi-agent problem, utilizing adversarial IRL to simultaneously recover the underlying reward functions of interacting vehicles and generate realistic trajectories.
Ref.~\cite{liu2024sociality} developed a game-theoretic IRL approach to quantify diverse social patterns and coupled decision-making during complex driving interactions. 
Ref.~\cite{li2025responsibility} proposed a hierarchical MAIRL framework that incorporates the concept of driving responsibility to capture socially compatible interactive behaviors. 
These studies highlight the need to extend driving reward learning from isolated agents to interaction-aware multi-agent settings. 
Nevertheless, this extension remains computationally challenging, as the joint action space grows exponentially with the number of interacting vehicles, making direct joint-policy inference intractable in realistic traffic scenarios.

Moreover, rewards learned solely from trajectory demonstrations may not necessarily reflect the true objectives of human drivers.  
Even when the learned rewards generate behaviorally plausible trajectories, they may primarily reflect statistical regularities in observed data rather than the subjective considerations behind human driving decisions. 
One potential solution is to incorporate human feedback signals, which are increasingly available in various forms (e.g., emotion, verbal comments, satisfaction ratings), as an additional source of information for interpreting and constraining the learned rewards.
Such signals can reveal drivers' relative preferences across different traffic states, actions, or interaction outcomes, thereby providing qualitative guidance on which situations should be assigned higher or lower rewards.
However, little attention has been paid to incorporating human feedback signals into the learning of human driving rewards. 

\noindent
\emph{Statement of Contributions:}
To address the aforementioned gaps, we propose a Human-Feedback-Guided Multi-Agent Inverse Reinforcement Learning (HFG-MAIRL) framework for interactive driving behavior modeling. Our contribution is two-fold. 
First, we develop a MAIRL framework that recovers interaction-aware reward functions from coupled multi-vehicle driving demonstrations. 
Unlike single-agent formulations that treat surrounding vehicles as part of the environment, our framework models vehicles as strategic agents whose decisions are mutually influenced through their interactions.
To enable tractable reward learning, we explicitly model each vehicle and its directly interacting neighbors while representing broader traffic conditions through aggregated traffic states. 
Second, to enhance the behavioral relevance of the learned rewards, we develop a human-feedback-guided component that incorporates available human feedback as ordinal preference supervision through a Bradley–Terry likelihood, thereby mitigating the ambiguity of trajectory-only reward learning and better aligning the learned rewards with drivers' subjective responses to traffic interactions. In this paper, we use driver emotion as an example instantiation of such human feedback. To the best of our knowledge, this is among the first attempts to incorporate human feedback signals into driving reward learning. 

The remainder of this paper is organized as follows. 
Section~\ref{sec:problem_formulation} presents the problem formulation. 
Section~\ref{sec:methodology} details the methodology of the proposed HFG-MAIRL framework.
Section~\ref{sec:result} evaluates the proposed framework on a multimodal driving dataset. 
Section~\ref{sec:conclusion} concludes the paper and presents future work. 

\section{Problem Formulation}
\label{sec:problem_formulation}

We consider an interactive driving scene represented by an interaction graph
$\mathcal{G}=(\mathcal{V},\mathcal{E})$, where $\mathcal{V}$ is the set of vehicles and $\mathcal{E}$ is the set of pairwise interaction edges.
An edge $(i,j)\in\mathcal{E}$ indicates that vehicle $j$ may influence the subsequent motion of vehicle $i$, such as when the two vehicles are successive in the same lane or adjacent in neighboring lanes.
The set of direct neighbors of vehicle $i$ is defined as
$
\mathcal{N}_i=\{j\in\mathcal{V}\mid(i,j)\in\mathcal{E}\}.
$
Here, we adopt the realistic assumption that the interaction topology evolves slowly over time.
For notational simplicity, the explicit time indices of $\mathcal{G}$, $\mathcal{E}$, and $\mathcal{N}_i$ are omitted when no ambiguity arises.

The driving process is discretized into time steps indexed by~$t$, with a fixed sampling interval $\Delta t$.
At time step $t$, the state of vehicle $i\in\mathcal{V}$ is
$x_{i,t}=(p_{x,i,t}, \allowbreak p_{y,i,t}, \allowbreak v_{i,t}, \allowbreak \psi_{i,t}, \allowbreak a_{i,t}, \allowbreak \omega_{i,t}),$
where $p_{x,i,t}$ and $p_{y,i,t}$ denote the longitudinal and lateral positions, $v_{i,t}$ is the longitudinal speed, $\psi_{i,t}$ is the heading angle, $a_{i,t}$ is the longitudinal acceleration, and $\omega_{i,t}$ is the yaw rate.
The joint state of all vehicles is denoted by
$
\boldsymbol{x}_t=\{x_{i,t}\}_{i\in\mathcal{V}}.
$

We define a sequence of decision epochs indexed by $\kappa$, with consecutive epochs separated by $H_{\mathrm{dec}}\Delta t$.
Let $n_\kappa$ denote the time step corresponding to the beginning of decision epoch $\kappa$.
The corresponding decision interval is $[n_\kappa,n_{\kappa+1})$, where
$
n_{\kappa+1}=n_\kappa+H_{\mathrm{dec}}.
$
At decision epoch $\kappa$, vehicles observe historical interaction states and select actions to be executed over the subsequent decision interval.
The historical states of all vehicles in the scene constitute a global historical interaction context:
$
\boldsymbol{\mathcal{X}}_{\kappa}
=
\{\boldsymbol{x}_{n_\kappa-h}\}_{h=0}^{H_{\mathrm{obs}}}.
$

During a driving episode, a driver monitoring system may provide interval-level human feedback annotations derived from in-cabin signals, such as video recordings of the driver's facial expressions, to characterize the driver's subjective response (e.g., emotion).
Let $\mathcal{V}_{\mathrm{fb}}\subseteq\mathcal{V}$ denote the subset of vehicles for which such feedback annotations are available.
For vehicle $i\in\mathcal{V}_{\mathrm{fb}}$, the feedback associated with decision interval $\kappa$ is represented by a categorical label $e_{i,\kappa}\in\mathcal{H}_{\mathrm{fb}}$, where $\mathcal{H}_{\mathrm{fb}}$ is a finite set of feedback categories.
As decision intervals are short relative to changes in the driver’s subjective state, such feedback labels are assumed to remain constant within an interval but may vary across intervals.

The goal of this study is to recover interaction-aware reward functions that rationalize real-world multi-vehicle interactions, based on demonstrations that include observed interaction contexts, joint actions, and available human feedback. Specifically, we define the demonstration dataset as
$
\mathcal{D}
=
\left\{
\left(
\zeta^{(m)},
\mathcal{E}_{\mathrm{fb}}^{(m)}
\right)
\right\}_{m=1}^{M},
$
where $M$ is the number of demonstration episodes indexed by $m$. In each episode $m$, $T_m$ is the number of decision intervals, $\mathcal{E}_{\mathrm{fb}}^{(m)}$ is the available human feedback annotations denoted by $
\mathcal{E}_{\mathrm{fb}}^{(m)}
=
\left\{
e_{i,\kappa}^{(m)}
\;\middle|\;
i\in\mathcal{V}_{\mathrm{fb}}^{(m)},\;
\kappa=0,\ldots,T_m-1
\right\}$, and $
\zeta^{(m)}
=
\left(
\left\{
\boldsymbol{\mathcal{X}}_{\kappa}^{(m)},
\boldsymbol{a}_{\kappa}^{*(m)}
\right\}_{\kappa=0}^{T_m-1},
\boldsymbol{\mathcal{X}}_{T_m}^{(m)}
\right)
$ contains the historical interaction contexts $\boldsymbol{\mathcal{X}}_{\kappa}^{(m)}$ and demonstrated joint decisions 
$\boldsymbol{a}_{\kappa}^{*(m)}$ in all 
decision epochs in episode $m$. Our proposed methodology is presented next. 
 
\section{Methodology} 
\label{sec:methodology}

\begin{figure*}
    \centering
    \begin{subfigure}[b]{0.98\textwidth}
        \includegraphics[width=0.98\textwidth]{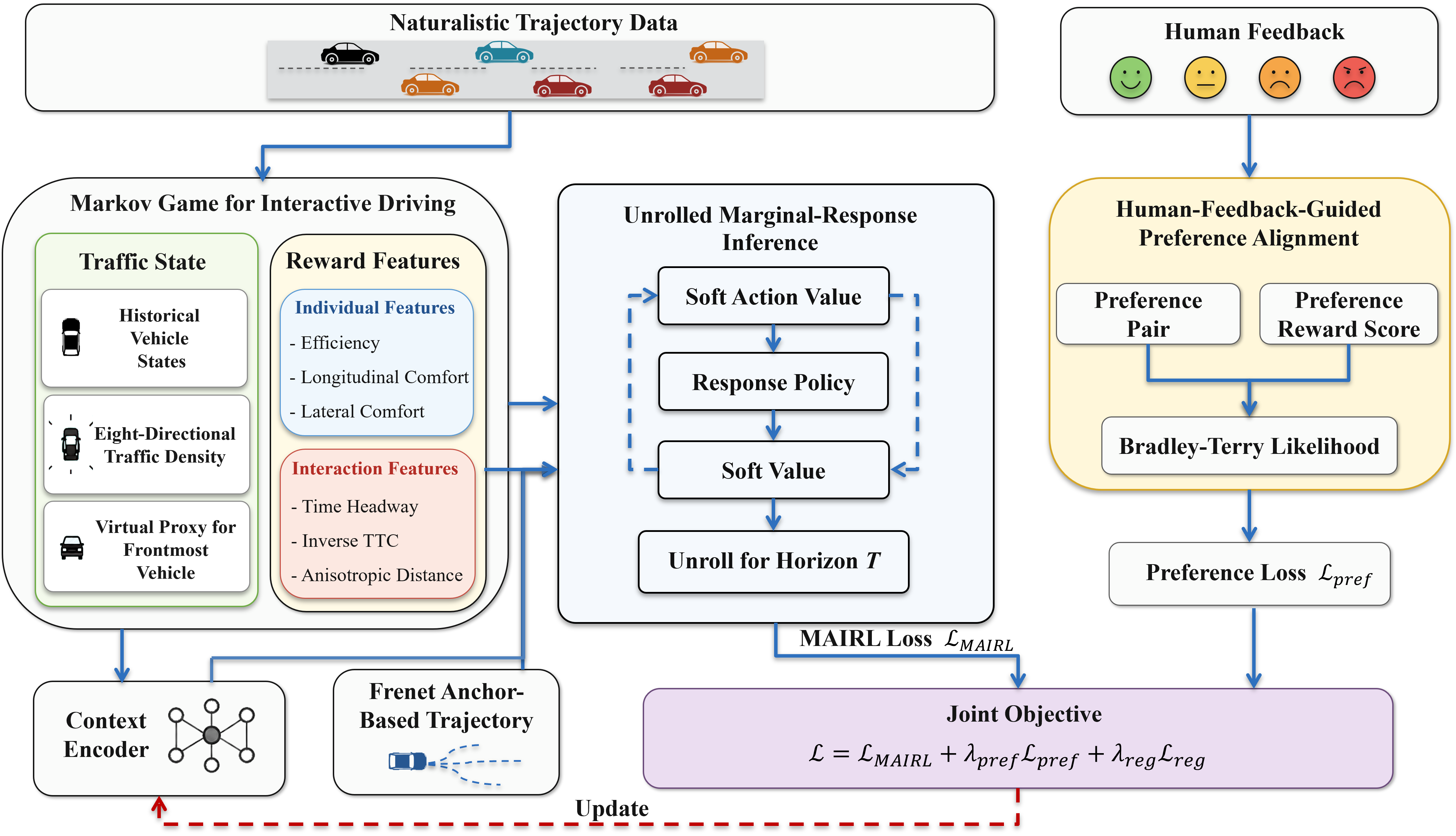}
    \end{subfigure}
    \caption{Diagram of the proposed human-feedback-guided MAIRL framework for interactive driving behavior modeling.}
    \label{fig:diagram}
\end{figure*}

As illustrated in Fig.~\ref{fig:diagram}, we propose a Human-Feedback-Guided Multi-Agent Inverse Reinforcement Learning (HFG-MAIRL) framework for interactive driving behavior modeling through reward learning from multi-vehicle demonstrations and human feedback. Section~\ref{ssec:meanfield} formulates the multi-vehicle driving process as a finite-horizon Markov game with interaction-aware reward functions. Building on this formulation, Section~\ref{ssec:mairl} develops a maximum-entropy-based MAIRL approach that learns context-dependent reward weights from demonstrated multi-vehicle decisions through iterative marginal-response inference. To make MAIRL tractable, we localize the interaction graph to each focal vehicle and its directly interacting neighbors, while the influence of vehicles outside the local graph is retained from aggregated traffic states. Section~\ref{sec:hitl} incorporates available human feedback as ordinal preference supervision to constrain the relative rewards of the corresponding demonstrated decisions.  

\subsection{Markov Game Formulation for Interactive Driving}
\label{ssec:meanfield}

We formulate the sequential interaction among vehicles as a finite-horizon Markov game, in which each vehicle selects a motion decision while accounting for the simultaneous decisions of surrounding vehicles. The Markov game is denoted by 
$
\mathfrak{M}
=
(
\mathcal{V},
\mathcal{S},\allowbreak
\{\mathcal{O}_i\}_{i\in\mathcal{V}},
\{\mathcal{A}_i\}_{i\in\mathcal{V}},
\mathcal{P},
\{r_i\}_{i\in\mathcal{V}},
\rho_0,
\gamma,
T
),
$
where $\mathcal{V}$ is the set of vehicles, $\mathcal{S}$ is the global decision-state space, $\mathcal{O}_i$ is the observation function of vehicle $i$, $\mathcal{A}_i$ is its action space, $\mathcal{P}$ is the transition kernel, $r_i$ is its reward function, $\rho_0$ is the initial-state distribution, $\gamma\in[0,1]$ is the discount factor, and $T$ is the episode horizon.

\noindent\emph{State.}
At decision epoch $\kappa$, the global state is defined by the historical interaction context introduced in Section~\ref{sec:problem_formulation}, i.e.,
$s_\kappa=\boldsymbol{\mathcal{X}}_\kappa$, which contains the historical states of all vehicles over the observation horizon.

\noindent\emph{Observation.} The observation of each vehicle is represented by $
o_{i,\kappa}
=
\mathcal{O}_i(s_\kappa)
=
\left(
\boldsymbol{\mathcal{X}}_{i,\kappa},
\boldsymbol{\mathcal{X}}_{\mathcal{N}_i,\kappa}^{\mathrm{rel}}
\right)
$, 
consisting of the vehicle's own state history $\boldsymbol{\mathcal{X}}_{i,\kappa}
=
\{x_{i,n_\kappa-h}\}_{h=0}^{H_{\mathrm{obs}}}$ and the relative state histories $\boldsymbol{\mathcal{X}}_{\mathcal{N}_i,\kappa}^{\mathrm{rel}} 
=
\left\{
\operatorname{Rel}(x_{j,n_\kappa-h},x_{i,n_\kappa-h})
\mid
j\in\mathcal{N}_i,\,
h=0,\ldots,H_{\mathrm{obs}}
\right\}$ of its directly interacting neighbors, where $\operatorname{Rel}(\cdot,\cdot)$ is the state of neighboring vehicles relative to vehicle $i$.

\noindent\emph{Action.}
At decision epoch $\kappa$, vehicle $i$ selects an action 
$a_{i,\kappa}\in\mathcal{A}_i(s_\kappa)=\{1,\ldots,K\}$ corresponding to a candidate trajectory
$\boldsymbol{\tau}_{i,\kappa}=F_i(s_\kappa,a_{i,\kappa})$, where $F_i(\cdot)$ denotes the trajectory generation function that maps the current decision state and selected action to a trajectory over the subsequent decision interval. 
The resulting candidate trajectory set is $\Omega_{i,\kappa}=\{F_i(s_\kappa,k)\}_{k=1}^{K}$, and the joint action space at decision epoch $\kappa$ is $\mathcal{A}_\kappa=\prod_{i\in\mathcal{V}}\mathcal{A}_i(s_\kappa)$.

The trajectory generation function $F_i(s_\kappa,k)$ maps action index $k$ to its corresponding terminal anchor $(\Delta s_k,\Delta d_k)$ and generates the associated trajectory according to the current state of vehicle $i$. 
For action index $k\in\{1,\ldots,K\}$, the terminal anchor $(\Delta s_k,\Delta d_k)$ is the $k$-th element obtained by uniformly discretizing the prescribed longitudinal and lateral displacement ranges $[\Delta s_{\min},\Delta s_{\max}]$ and $[\Delta d_{\min},\Delta d_{\max}]$. Thus, the same action index corresponds to the same terminal-anchor specification across vehicles. 
Given the selected anchor, the corresponding trajectory is generated in the Frenet frame using quintic polynomials $s_k(\xi)=\sum_{r=0}^{5}c_r\xi^r$ and $d_k(\xi)=\sum_{r=0}^{5}l_r\xi^r$, where $\xi\in[0,H_{\mathrm{dec}}\Delta t]$ denotes the local continuous time within the decision interval. 
The polynomial coefficients are determined by the current vehicle state, the selected terminal anchor, and the endpoint kinematic conditions. 
Specifically, to ensure continuity across consecutive intervals, the longitudinal and lateral trajectories in the Frenet frame at the beginning of each interval ($\xi=0$) match the positions, velocities, and accelerations at the end of the preceding interval.
After discretization, the remaining kinematic states are obtained from the generated trajectory, yielding a finite action space. The joint action $\boldsymbol{a}_\kappa=\{a_{i,\kappa}\}_{i\in\mathcal{V}}$ determines the within-interval motion of all vehicles. 

\noindent\emph{Transition.} After executing the actions over the decision interval $[n_\kappa,n_{\kappa+1})$, the next state is obtained by rolling out the vehicle kinematics according to $s_{\kappa+1}=G(s_\kappa,\boldsymbol{a}_\kappa)$, where $G(\cdot)$ denotes the trajectory-rollout transition that maps the current state and joint action to the state at the next decision epoch.

\noindent\emph{Reward.} The reward function of each vehicle is represented by
\begin{equation}
r_i(s_\kappa,\boldsymbol{a}_\kappa)
=
\boldsymbol{\theta}_{\mathrm{ind},i,\kappa}^{\top}
\widetilde{\boldsymbol{\phi}}_{\mathrm{ind}}
\left(
\boldsymbol{\tau}_{i}
\right)
+
\boldsymbol{\theta}_{\mathrm{int},i,\kappa}^{\top}
\widetilde{\boldsymbol{\Phi}}_{\mathrm{int},i}(\boldsymbol{\tau}_{i}, \boldsymbol{\tau}_{j}),
\label{eq:reward_main}
\end{equation}
where $\widetilde{\boldsymbol{\phi}}_{\mathrm{ind}}$ and $\widetilde{\boldsymbol{\Phi}}_{\mathrm{int},i}$ denote the normalized individual and aggregated interaction feature vectors, respectively. Their raw feature representations are defined as follows:
\begin{equation}
\begin{aligned}
\boldsymbol{\phi}_{\mathrm{ind}}(\boldsymbol{\tau}_i)
&=
\left[
\phi_{\mathrm{eff}}(\boldsymbol{\tau}_i),
\phi_{\mathrm{comf}}^{\mathrm{lon}}(\boldsymbol{\tau}_i),
\phi_{\mathrm{comf}}^{\mathrm{lat}}(\boldsymbol{\tau}_i)
\right]^{\top},
\\
\boldsymbol{\phi}_{\mathrm{int}}(\boldsymbol{\tau}_i,\boldsymbol{\tau}_j)
&=
\left[
\phi_{\mathrm{THW}}(\boldsymbol{\tau}_i,\boldsymbol{\tau}_j),
\phi_{\mathrm{InvTTC}}(\boldsymbol{\tau}_i,\boldsymbol{\tau}_j),
\phi_{\mathrm{ell}}(\boldsymbol{\tau}_i,\boldsymbol{\tau}_j)
\right]^{\top}.
\end{aligned}
\label{eq:reward_feature_vectors}
\end{equation}
The individual features describe the efficiency and comfort of vehicle $i$, while the interaction features characterize its safety-related and spatial interactions with vehicle $j$.
Here, the interactive feature vector in Eq.~\eqref{eq:reward_main} aggregates the pairwise interaction features between vehicle $i$ and all its directly interacting neighbors as $\boldsymbol{\Phi}_{\mathrm{int},i}(\boldsymbol{\tau}_{i},\boldsymbol{\tau}_{j})=\sum_{j\in\mathcal{N}_i}\boldsymbol{\phi}_{\mathrm{int}}(\boldsymbol{\tau}_{i},\boldsymbol{\tau}_{j})$, where the second argument of $\boldsymbol{\Phi}_{\mathrm{int},i}$ collectively represents the trajectories of the directly interacting neighbors in $\mathcal{N}_i$.
Their corresponding context-dependent weights are learned in Section~\ref{ssec:mairl}. Below, we introduce each term of the features. 

\vspace{0.2em} \noindent\emph{(i) Efficiency.}
The efficiency feature penalizes deviation from a reference speed:
\begin{equation}
\phi_{\mathrm{eff}}(\boldsymbol{\tau}_i)
=
-
\frac{1}{H_{\mathrm{dec}}+1}
\sum_{h=0}^{H_{\mathrm{dec}}}
\left(
\frac{v_{i,h}-v_{\mathrm{ref}}}{v_{\mathrm{ref}}}
\right)^2,
\label{eq:eff_feature}
\end{equation}
where $v_{i,h}$ is the longitudinal speed of vehicle $i$ at local step $h$, and $v_{\mathrm{ref}}$ is the reference speed, which is set to represent the typical driving speed of the considered traffic scene.

\vspace{0.2em} \noindent\emph{(ii) Comfort}. We consider both longitudinal and lateral comfort terms to ensure stable driving. The longitudinal term penalizes excessive acceleration:
\begin{equation}
\phi_{\mathrm{comf}}^{\mathrm{lon}}(\boldsymbol{\tau}_i)
=
-
\frac{1}{H_{\mathrm{dec}}+1}
\sum_{h=0}^{H_{\mathrm{dec}}}
\frac{|a_{i,h}|}{a_{\max}},
\label{eq:lon_comfort_feature}
\end{equation}
while the lateral comfort term penalizes excessive yaw rate:
\begin{equation}
\phi_{\mathrm{comf}}^{\mathrm{lat}}(\boldsymbol{\tau}_i)
=
-
\frac{1}{H_{\mathrm{dec}}+1}
\sum_{h=0}^{H_{\mathrm{dec}}}
\frac{|\omega_{i,h}|}{\omega_{\max}},
\label{eq:lat_comfort_feature}
\end{equation}
where $a_{\max}$ and $\omega_{\max}$ denote the prescribed maximum magnitudes of longitudinal acceleration and yaw rate, respectively, and are used for normalization.

\vspace{0.2em} \noindent\emph{(iii) Time headway.} The estimated time headway is penalized to enhance safety: 
\begin{equation}
\phi_{\mathrm{THW}}
(\boldsymbol{\tau}_i,\boldsymbol{\tau}_j)
=
-
\max_{h=0,\ldots,H_{\mathrm{dec}}}
w_{ij}^{\mathrm{front}}(h)
\frac{1}{\max(\mathrm{THW}_{ij}(h),\epsilon)},
\label{eq:thw_feature}
\end{equation}
where
$\mathrm{THW}_{ij}(h)
=
(p_{x,j,h}-p_{x,i,h})/\max(v_{i,h},\epsilon)$ estimates the time headway as the ratio between forward spacing and speed. Binary variable $w_{ij}^{\mathrm{front}}(h)\in\{0,1\}$ indicates whether  vehicle $j$ is a same-lane leader of vehicle $i$. A small $\epsilon>0$ is used to avoid numerical instability.

\vspace{0.2em} \noindent\emph{(iv) Inverse time-to-collision (TTC).}
The inverse-TTC captures imminent same-lane longitudinal collision risk:
\begin{equation}
\phi_{\mathrm{InvTTC}}
(\boldsymbol{\tau}_i,\boldsymbol{\tau}_j)
=
-
\max_{h=0,\ldots,H_{\mathrm{dec}}}
w_{ij}^{\mathrm{front}}(h)
\frac{1}{\mathrm{TTC}_{ij}(h)+\epsilon},
\label{eq:invttc_feature}
\end{equation}
where
$\mathrm{TTC}_{ij}(h)
=
(p_{x,j,h}-p_{x,i,h})/(v_{i,h}-v_{j,h}+\epsilon)$
when $w_{ij}^{\mathrm{front}}(h)=1$ and $v_{i,h}>v_{j,h}$, and is set to $+\infty$ otherwise.

\vspace{0.2em}\noindent\emph{(v) Anisotropic social distance.}
The anisotropic social-distance feature captures directional spatial proximity between interacting vehicles:
\begin{equation}
\phi_{\mathrm{ell}}
(\boldsymbol{\tau}_i,\boldsymbol{\tau}_j)
=
-
\max_{h=0,\ldots,H_{\mathrm{dec}}}
\exp
\left(
-\frac{(\Delta p_{x,h})^2}{2\sigma_x^2}
-\frac{(\Delta p_{y,h})^2}{2\sigma_y^2}
\right).
\label{eq:ell_feature}
\end{equation}
where $\Delta p_{x,h}=p_{x,j,h}-p_{x,i,h}$ and $\Delta p_{y,h}=p_{y,j,h}-p_{y,i,h}$ denote the longitudinal and lateral relative displacements of vehicle $j$ with respect to vehicle $i$, respectively. We set $\sigma_x>\sigma_y$ to impose greater sensitivity to lateral proximity.

Following~\cite{huang2021driving}, we normalize all six reward features after aggregating the pairwise interaction features. Each feature is normalized as $\widetilde{f}_j=f_j/\alpha_j$, where $\alpha_j$ is the 95th percentile of nonzero absolute values of the corresponding feature in the training set. The normalization scales remain fixed throughout training and evaluation.

\noindent \emph{Policy objective.} Under a joint policy profile $\boldsymbol{\pi} = (\boldsymbol{\pi}_i, \boldsymbol{\pi}_{-i})$ with $-i$ indicating all other vehicles, the expected discounted return of vehicle $i$ over a finite horizon is 
\begin{equation}
J_i(\boldsymbol{\pi}_i, \boldsymbol{\pi}_{-i})
=
\mathbb{E}_{\rho_0,\mathcal{P},\boldsymbol{\pi}_i, \boldsymbol{\pi}_{-i}}
\left[
\sum_{\kappa=0}^{T-1}
\gamma^\kappa
r_i(s_\kappa,\boldsymbol{a}_\kappa)
\right].
\label{eq:vehicle_expected_return}
\end{equation}

Given the policies of the other vehicles $\boldsymbol{\pi}_{-i}$, vehicle $i$ seeks to optimize its own policy: 
\begin{align}
    \boldsymbol{\pi}_i^* \in \arg\max_{\boldsymbol{\pi}_i} J_i(\boldsymbol{\pi}_i,\boldsymbol{\pi}_{-i}).
\end{align}
Because the reward and state transition depend on the joint actions, the optimal response of each vehicle generally depends on the policies of the other interacting vehicles. The resulting Markov game therefore provides a forward model of interactive driving decisions, which will subsequently be used to infer the context-dependent reward weights from demonstrations.

\subsection{Localized Multi-Agent Inverse Reinforcement Learning}
\label{ssec:mairl}

With the Markov game formulated above, this subsection leverages MAIRL to infer the parameters $\boldsymbol{\theta}_{\mathrm{int},i,\kappa}$ and $\boldsymbol{\theta}_{\mathrm{ind},i,\kappa}$ in Eq.~\eqref{eq:reward_main} from interactive driving demonstrations, which characterize driver preferences over different performance criteria. 
However, directly applying MAIRL in typical traffic scenarios may require a large interaction graph. In particular, while a vehicle's decision depends on its neighboring vehicles, those neighbors' decisions are also influenced by their own neighbors. Accounting for such dependencies therefore requires the interaction graph to extend beyond the immediate neighborhood and, recursively, across a potentially large portion of the traffic system. Explicitly modeling such a large interaction graph poses two challenges. First, it requires demonstration datasets covering a correspondingly large portion of the traffic system, which are typically unavailable. Second, it substantially increases the computational complexity of solving the Markov game. 

To address these challenges, we construct a local interaction graph for each vehicle that explicitly includes only the vehicle itself and its immediate neighbors. This confines both the Markov game computation and the required demonstration data to a small local graph. To retain the influence of vehicles outside this graph, we represent their collective effects on each neighboring vehicle through aggregated traffic states, analogous to mean-field states. This localized representation therefore captures the influence of the broader traffic environment while substantially reducing the required data coverage and computational complexity.

Accordingly, the presentation of our methodology is divided into two parts. We first present the MAIRL formulation for the general Markov game, and then describe how it is applied to traffic scenarios through a localized Markov game with aggregated traffic states.

\subsubsection{Maximum-Entropy-Based MAIRL}
\label{sssec:MAIRL}
Given the Markov-game formulation above, we use MAIRL to infer the reward weights from interactive driving demonstrations. Following the maximum-entropy principle, the inverse problem is formulated probabilistically. Rather than assuming that the demonstrated actions are strictly optimal, actions with higher expected returns are assigned higher probabilities, and the reward weights are learned by maximizing the likelihood of the observed actions. 

Instead of independently estimating the reward weights for each vehicle at every decision epoch, we generate them from the vehicle observation using a shared context encoder $f_{\mathrm{ctx}}
(\cdot)$ with learnable parameters $\boldsymbol{\eta}$. For vehicle $i$ at decision epoch $\kappa$, its reward weights are generated by 
\begin{equation}
\boldsymbol{\theta}_{i,\kappa}(\boldsymbol{\eta})
=
\operatorname{softplus}
\left(
f_{\mathrm{ctx}}
\left(
o_{i,\kappa};
\boldsymbol{\eta}
\right)
\right), 
\label{eq:context_encoder}
\end{equation}
where
$\boldsymbol{\theta}_{i,\kappa}
=
[
\boldsymbol{\theta}_{\mathrm{ind},i,\kappa}^{\top},
\boldsymbol{\theta}_{\mathrm{int},i,\kappa}^{\top}
]^{\top}$,
with
$\boldsymbol{\theta}_{\mathrm{ind},i,\kappa}
=
[
\theta_{\mathrm{eff},i,\kappa},
\theta_{\mathrm{comf},i,\kappa}^{\mathrm{lon}},
\theta_{\mathrm{comf},i,\kappa}^{\mathrm{lat}}
]^{\top}$
and
$\boldsymbol{\theta}_{\mathrm{int},i,\kappa}
=
[
\theta_{\mathrm{THW},i,\kappa},
\theta_{\mathrm{InvTTC},i,\kappa},
\theta_{\mathrm{ell},i,\kappa}
]^{\top}$.
Note that sharing the encoder across vehicles allows demonstrations from different vehicles to jointly inform preference learning, while conditioning on $o_{i,\kappa}$ allows the inferred reward weights to vary across vehicles and traffic contexts.

For a given reward model, evaluating the likelihood of a demonstrated action requires computing the joint policy $\boldsymbol{\pi}_{\kappa}(\boldsymbol{a}_\kappa\mid\boldsymbol{o}_\kappa)$ at each epoch $\kappa$, which can be computationally expensive, as the size of the joint action space grows exponentially with the number of interacting vehicles. 
To reduce the computational burden, we approximate the joint policy using vehicle-level marginal policies, where each vehicle maintains its own policies and iteratively responds to the policies of others (see, e.g.,~\cite{zhang2024gradient}):
\begin{equation}
\boldsymbol{\pi}_{\kappa}\left(\boldsymbol{a}_\kappa\mid\boldsymbol{o}_\kappa\right)\approx\prod_{i\in\mathcal{V}}\pi_{i,\kappa}\left(a_i\mid o_{i,\kappa}\right),\qquad \boldsymbol{o}_\kappa=\{o_{i,\kappa}\}_{i\in\mathcal{V}}.
\label{eq:factorized_joint_policy}
\end{equation}
It is worth noting that although Eq.~\eqref{eq:factorized_joint_policy} factorizes the joint policy, it does not remove the coupling among vehicle decisions. Each marginal policy is updated in response to the current marginal policies of the other vehicles, while their interactions remain captured through the reward and transition dynamics. Iterating these updates therefore allows changes in one vehicle's policy to affect the policies of the others without explicitly computing the full joint policy. 

At marginal-response iteration $q$, vehicle $i$ evaluates each candidate action against the current marginal policies of the other vehicles, i.e., 
\begin{equation}
\begin{aligned}
Q_{i,\kappa}^{(q)}(a_i)
&=
\mathbb{E}_{
\substack{
\boldsymbol{a}_{-i,\kappa}
\sim
\prod_{j\in\mathcal{V}\setminus\{i\}}
\pi_{j,\kappa}^{(q)}
(\cdot\mid o_{j,\kappa})
}
}
\Bigg[
r_i
\left(
s_\kappa,
(a_i,\boldsymbol{a}_{-i,\kappa})
\right)
\\
&\qquad
+
\gamma
V_{i,\kappa+1}
\Bigg(
\mathcal{O}_i
\Big(
G
\big(
s_\kappa,
(a_i,\boldsymbol{a}_{-i,\kappa})
\big)
\Big)
\Bigg)
\Bigg].
\end{aligned}
\label{eq:explicit_soft_bellman}
\end{equation}
Here, $V_{i,\kappa+1}$ denotes the continuation value at the successor state generated by $G$ for each candidate joint action. It is computed recursively by generating candidate trajectories, updating context-dependent reward weights, and performing marginal-response inference at each successor state. The expectation can be estimated by sampling the other vehicles' actions from their factorized marginal policies, with terminal condition $V_{i,T_m}(\cdot)=0$ for episode $m$.

Vehicle $i$ then updates its marginal response according to
\begin{equation}
\pi_{i,\kappa}^{(q+1)}
\left(
a_i\mid o_{i,\kappa}
\right)
=
\frac{
\exp
\left(
\beta Q_{i,\kappa}^{(q)}(a_i)
\right)
}{
\displaystyle
\sum_{\widetilde a_i\in\mathcal{A}_i(s_\kappa)}
\exp
\left(
\beta Q_{i,\kappa}^{(q)}(\widetilde a_i)
\right)
},
\label{eq:explicit_logit_response}
\end{equation}
where $\beta>0$ is the inverse-temperature parameter.
The corresponding soft value is
\begin{equation}
V_{i,\kappa}^{(q+1)}(o_{i,\kappa})
=
\frac{1}{\beta}
\log
\sum_{a_i\in\mathcal{A}_i(s_\kappa)}
\exp
\left(
\beta Q_{i,\kappa}^{(q)}(a_i)
\right).
\label{eq:explicit_soft_value}
\end{equation}

For a finite episode, the value computation proceeds backward from $\kappa=T-1$ to $\kappa=0$, evaluating continuation values at the successor states induced by candidate joint actions. At each decision epoch, the continuation-value function $V_{i,\kappa+1}(\cdot)$ is held fixed while the coupled marginal policies are approximated by unrolling $N_{\mathrm{resp}}$ marginal-response iterations. After $N_{\mathrm{resp}}$ iterations, the resulting action value, marginal policy, and soft value are referred to as the finalized response quantities and denoted by $Q_{i,\kappa}(a_i)$, $\pi_{i,\kappa}(a_i\mid o_{i,\kappa})$, and $V_{i,\kappa}(o_{i,\kappa})$, respectively.

The inverse-learning step then adjusts context-encoder parameters $\boldsymbol{\eta}$ by assigning high likelihood to the demonstrated multi-vehicle decisions under the finalized response distributions. For demonstration episode $m$, the demonstrated trajectory of vehicle $i$ at decision epoch $\kappa$ is matched to the candidate in $\Omega_{i,\kappa}$ with the closest terminal displacement $(\Delta s_{i,\kappa}^{*},\Delta d_{i,\kappa}^{*})$, and the corresponding candidate index is taken as the demonstrated action $a_{i,\kappa}^{*}$. 

Using the finalized action values as maximum-entropy decision scores, the demonstrated-action log-likelihood is
\begin{equation}
\log
\widetilde{\pi}_{i,\kappa}
\left(
a_{i,\kappa}^{*}
\mid
o_{i,\kappa}
\right)
=
\beta Q_{i,\kappa}(a_{i,\kappa}^{*})
-
\log
\sum_{a_i\in\mathcal{A}_{i,\kappa}}
\exp
\left(
\beta Q_{i,\kappa}(a_i)
\right).
\label{eq:expert_loglikelihood}
\end{equation}
The episode-level MAIRL loss is
\begin{equation}
\mathcal{L}_{\mathrm{MAIRL}}^{(m)}
=
-
\sum_{\kappa=0}^{T_m-1}
\sum_{i\in\mathcal{V}^{(m)}}
\mathbb{I}_{i,\kappa}^{(m)}
\log
\widetilde{\pi}_{i,\kappa}^{(m)}
\left(
a_{i,\kappa}^{*(m)}
\mid
o_{i,\kappa}^{(m)}
\right),
\label{eq:episode_mairl_loss}
\end{equation}
where $\mathbb{I}_{i,\kappa}^{(m)}$ indicates whether vehicle $i$ has a valid demonstrated action at decision epoch $\kappa$.
For a mini-batch $\mathcal{B}$, the MAIRL loss is
\begin{equation}
\mathcal{L}_{\mathrm{MAIRL}}
=
\frac{1}{|\mathcal{B}|}
\sum_{m\in\mathcal{B}}
\mathcal{L}_{\mathrm{MAIRL}}^{(m)}.
\label{eq:mairl_loss_final}
\end{equation}

\subsubsection{Localized Markov Game with Aggregated Traffic States}
\label{sssec:mean_filed_macro}
To make the Markov game tractable with locally available driving demonstrations, we localize the interaction graph by explicitly retaining only each focal vehicle and its directly interacting neighbors. Simply truncating the graph, however, would neglect the influence of vehicles outside the local graph, which can still affect the decisions of the explicitly represented vehicles through the surrounding traffic conditions. We therefore summarize these external vehicles using fixed-dimensional aggregated traffic states within the localized Markov game. Overall, such localization affects two components of the original Markov-game formulation: (i) the observation of each explicitly represented vehicle is augmented with the aggregated traffic state to retain the influence of the surrounding traffic, and (ii) the forward-safety interaction of a frontmost vehicle is evaluated against a virtual proxy constructed from the aggregated traffic state when no real leading vehicle is explicitly represented.

\vspace{0.2em} \noindent \emph{Aggregated traffic state}. The observation $o_{i,\kappa}$ is augmented with an aggregated traffic state $
\boldsymbol{\mathcal{H}}_{i,\kappa}^{\rho}
=
\left\{
\boldsymbol{\rho}_{i,n_\kappa-h}
\right\}_{h=0}^{H_{\mathrm{obs}}}
$, representing a short history of local traffic densities around vehicle $i$. 
Specifically, the surrounding region is partitioned into eight lane-aligned cells $\mathcal{D}_8=\{F,FL,L,BL,B,BR,R,FR\}$, corresponding to front, front-left, left, rear-left, rear, rear-right, right, and front-right cells around vehicle $i$, respectively. Each cell covers approximately one lane in width and 20--30 m longitudinally, and $\boldsymbol{\rho}_{i,t}^d$ denotes the traffic density within cell $d \in \mathcal{D}_8$ at time step $t$.
The cells are oriented according to the local direction of the lane occupied by vehicle $i$, and their densities are collected as $\boldsymbol{\rho}_{i,t}
=
[\rho_{i,t}^{F},
\allowbreak \rho_{i,t}^{FL},
\allowbreak \rho_{i,t}^{L},
\allowbreak \rho_{i,t}^{BL},
\allowbreak \rho_{i,t}^{B},
\allowbreak \rho_{i,t}^{BR},
\allowbreak \rho_{i,t}^{R},
\allowbreak \rho_{i,t}^{FR}]^{\top}$. 

This representation retains spatial coverage of the surrounding traffic while avoiding explicit modeling of every vehicle. Directly interacting neighbors are represented at the microscopic level, whereas vehicles beyond the local interaction graph are summarized at the macroscopic level through traffic density. The latter retain the aggregate traffic information within each local cell while sacrificing vehicle-level resolution, thereby providing complementary information about the broader traffic environment. Such local traffic densities can be obtained or estimated from onboard perception or other available traffic measurements.

\vspace{0.2em} \noindent \emph{Virtual proxy for forward-safety evaluation}. The aggregated traffic state also modifies the forward-safety evaluation for a frontmost vehicle in the local interaction graph. Although such a vehicle has no explicitly represented leader, traffic beyond the local graph can still constrain its forward motion. Since the time-headway and inverse-TTC features require a reference vehicle trajectory, we construct a virtual proxy from the aggregated traffic state as an equivalent leader.

For each nonempty lane $\ell$, let 
$f_{\ell,\kappa}$ denote its frontmost explicitly represented vehicle.
Its virtual proxy at decision epoch $\kappa$ is characterized by $
\boldsymbol{z}_{\ell,\kappa}^{\mathrm{vir}}
=
\left[
g_{\ell,\kappa}^{\mathrm{vir}},
v_{\ell,\kappa}^{\mathrm{vir}}
\right]^{\top}
$. The equivalent gap $g_{\ell,\kappa}^{\mathrm{vir}}$ is determined as the inverse of the local traffic density around $f_{\ell,\kappa}$, and clipped between physical bounds $g_{\min}$ and $g_{\max}$. The equivalent speed $v_{\ell,\kappa}^{\mathrm{vir}}$ is set to the local mean traffic speed. 
The virtual proxy is constructed at decision epoch $\kappa$ from the aggregated traffic state, and its trajectory is then generated over the current decision interval from its gap and speed. 

For the frontmost vehicle $f_{\ell,\kappa}$, the virtual proxy serves only as an equivalent leader for evaluating the forward-safety components of the interaction feature. Accordingly, 
\begin{equation}
\boldsymbol{\phi}_{\mathrm{int}}^{\mathrm{vir}}
\left(
\boldsymbol{\tau}_{i,\kappa},
\bar{\boldsymbol{\tau}}_{\ell,\kappa}^{\mathrm{vir}}
\right)
=
\left[
\phi_{\mathrm{THW}}
\left(
\boldsymbol{\tau}_{i,\kappa},
\bar{\boldsymbol{\tau}}_{\ell,\kappa}^{\mathrm{vir}}
\right),
\phi_{\mathrm{InvTTC}}
\left(
\boldsymbol{\tau}_{i,\kappa},
\bar{\boldsymbol{\tau}}_{\ell,\kappa}^{\mathrm{vir}}
\right),
0
\right]^{\top}.
\label{eq:virtual_proxy_interaction_feature}
\end{equation}
where the first two components are evaluated using the virtual-proxy trajectory in the same manner as for a real leading vehicle. The social-distance component is set to zero because the proxy represents aggregated traffic rather than a physical vehicle. The virtual proxy is generated exogenously from the aggregated traffic state and is therefore not treated as an agent in the Markov game, with no action, policy, reward, or demonstrated-action likelihood of its own.

\subsection{Human-Feedback-Guided Preference Alignment}
\label{sec:hitl}

The MAIRL approach developed above aims to infer reward weights from interactive driving demonstrations, but demonstrations alone only provide behavioral evidence and may not fully reveal the subjective preferences underlying the observed decisions.
For example, similar driving behaviors may lead to different subjective evaluations depending on the driver's perceived comfort, safety, or satisfaction.
To reduce this ambiguity, we incorporate human feedback, represented by interval-level labels characterizing the driver's subjective response to the experienced driving behavior (e.g., emotional responses), which is assumed to be available for a subset of vehicles $\mathcal{V}_{\mathrm{fb}}\subseteq\mathcal{V}$. 
Such feedback can be readily obtained from increasingly common in-cabin driver monitoring modalities, such as video recordings of facial expressions, and subsequently processed or annotated into feedback labels. By providing subjective evaluations associated with demonstrated state-action pairs, human feedback offers complementary supervision for aligning the learned reward with human preferences.

Human feedback is associated with each demonstration instance $\chi=(m,\kappa)$, where $m$ indexes the demonstration episode and $\kappa$ indexes the decision interval within that episode. Following the definitions in the Markov game, the corresponding instance contains the global interaction state $s^\chi = s_\kappa^{(m)}$, the vehicle-specific observation $o_i^\chi = \mathcal{O}_i(s^\chi)$, the demonstrated joint decision $\boldsymbol{a}^{*,\chi} = \boldsymbol{a}_\kappa^{*(m)}$ with corresponding interval-level trajectory $\boldsymbol{\tau}_i^{*,\chi} = F_i(s^\chi,a_i^{*,\chi})$, and the available human feedback label $e_i^\chi = e_{i,\kappa}^{(m)}$ for each feedback-available vehicle $i\in\mathcal{V}_{\mathrm{fb}}$. 

Instead of directly converting categorical human feedback into numerical reward values, we use it only to establish preference orderings over the corresponding demonstrated state-action pairs. 
To this end, we introduce a task-specific scoring function $
S:\mathcal{H}_{\mathrm{fb}}\rightarrow\mathbb{R}
$ to assign an ordinal preference level to each feedback category. For two instances $\chi^+=(m^+,\kappa^+)$ and $\chi^-=(m^-,\kappa^-)$, the feedback indicates that $\chi^+$ is preferred to $\chi^-$ by driver $i$ if $S(e_i^{\chi^+})>S(e_i^{\chi^-})$. Feedback categories with the same score are treated as ties and do not form preference pairs. 
Accordingly, the preference set for feedback-providing vehicle $i$ is
\begin{equation}
\mathcal{C}_{i,\mathrm{pref}}
=
\left\{
(\chi^+,\chi^-)
\;\middle|\;
S(e_i^{\chi^+})>S(e_i^{\chi^-})
\right\}.
\label{eq:preference_pair_set}
\end{equation}
This construction uses the feedback only to determine relative preference and does not impose arbitrary cardinal differences between feedback categories.

Since each feedback label characterizes the subjective response to an experienced state-action pair, we evaluate the learned reward directly on the corresponding demonstrated instance as $U_i^\chi=r_i\left(s^\chi,\boldsymbol{a}^{*,\chi}\right)$. 
Human feedback is then used to constrain the relative ordering of these rewards rather than their absolute values.
Specifically, for each preference pair $(\chi^+,\chi^-)\in\mathcal{C}_{i,\mathrm{pref}}$, the learned reward is expected to satisfy $U_i^{\chi^+}>U_i^{\chi^-}$. 

To incorporate this ordinal supervision into learning, we adopt a Bradley--Terry model, under which the probability that the learned reward assigns the feedback-consistent ordering is
\begin{equation}
P_i \left( (\chi^+, \chi^-) \in  \mathcal{C}_{i,\mathrm{pref}}\right)
= \sigma \left( U_i^{\chi^+} - U_i^{\chi^-} \right), i \in \mathcal{V}_{\mathrm{fb}}
\label{eq:bt_likelihood}
\end{equation}
where $\sigma(x)=1/(1+\exp(-x))$ is the sigmoid function. 

For each mini-batch containing at least one preference pair, the preference loss for vehicle $i$ is defined as
\begin{equation}
\begin{aligned}
\mathcal{L}_{i,\mathrm{pref}} =
-\frac{1}{|\mathcal{C}_{i,\mathrm{pref}}^{\mathcal{B}}|}
\sum_{(\chi^+,\chi^-)\in
\mathcal{C}_{i,\mathrm{pref}}^{\mathcal{B}}}
\log \sigma
\left(
U_i^{\chi^+}-U_i^{\chi^-}
\right).
\end{aligned}
\label{eq:pref_loss_agent}
\end{equation}
Here, $\mathcal{C}_{i,\mathrm{pref}}^{\mathcal{B}}$ denotes the preference pairs of vehicle $i$ constructed from the decision instances available in mini-batch $\mathcal{B}$. If no preference pair is available, $\mathcal{L}_{i,\mathrm{pref}}$ is set to zero.

The total preference loss over all feedback-providing vehicles in the mini-batch is
\begin{equation}
\mathcal{L}_{\mathrm{pref}}
=
\sum_{i\in\mathcal{V}_{\mathrm{fb}}}
\mathcal{L}_{i,\mathrm{pref}}.
\label{eq:pref_loss}
\end{equation}

Since the preference likelihood is determined by reward margins, excessively scaling the reward weights can increase the confidence of the inferred ordering without providing additional preference information. We therefore regularize the magnitude of the learned reward weights,
\begin{equation}
\mathcal{L}_{\mathrm{reg}}
=
\frac{1}{|\mathcal{B}|}
\sum_{m\in\mathcal{B}}
\sum_{\kappa=0}^{T_m-1}
\frac{
\displaystyle
\sum_{i\in\mathcal{V}^{(m)}}
\mathbb{I}_{i,\kappa}^{(m)}
\left\|
\boldsymbol{\theta}_{i,\kappa}^{(m)}
\right\|_2^2
}{
\displaystyle
\sum_{i\in\mathcal{V}^{(m)}}
\mathbb{I}_{i,\kappa}^{(m)}
+
\epsilon
}.
\label{eq:reg_loss}
\end{equation}

The final training objective combines behavioral inference from multi-agent demonstrations with preference alignment from human feedback, i.e., 
\begin{equation}
\mathcal{L} =
\mathcal{L}_{\mathrm{MAIRL}} + \lambda_{\mathrm{pref}}
\mathcal{L}_{\mathrm{pref}} + \lambda_{\mathrm{reg}} \mathcal{L}_{\mathrm{reg}}.
\label{eq:joint_objective}
\end{equation}
where the MAIRL term $\mathcal{L}_{\mathrm{MAIRL}}$ encourages the learned reward to explain observed driving decisions, while the preference term $\mathcal{L}_{\mathrm{pref}}$ further constrains its ordering using subjective human feedback. Their combination allows the reward model to remain behaviorally grounded while better reflecting human preferences.

The overall training procedure of the proposed HFG-MAIRL framework is summarized in Algorithm~\ref{alg:hfg_mairl}.

\begin{algorithm}[t]
\caption{Training Procedure of HFG-MAIRL}
\label{alg:hfg_mairl}
\small
\begin{algorithmic}[1]

\REQUIRE \parbox[t]{0.82\linewidth}{
Demonstration dataset $\mathcal{D}$,
candidate number $K$,
feedback scoring function $S(\cdot)$,
loss weights $\lambda_{\mathrm{pref}}$ and $\lambda_{\mathrm{reg}}$,
response iterations $N_{\mathrm{resp}}$,
and training epochs $N_{\mathrm{epoch}}$
}

\ENSURE Learned context-encoder parameters $\boldsymbol{\eta}$

\STATE Estimate and fix the normalization scales $\{\alpha_j\}_{j=1}^{6}$ from the training data

\STATE Initialize $\boldsymbol{\eta}$

\FOR{$epoch=1$ to $N_{\mathrm{epoch}}$}
    \FOR{each mini-batch $\mathcal{B}\subset\mathcal{D}$}

        \FOR{each episode $m\in\mathcal{B}$}

            \STATE \parbox[t]{0.82\linewidth}{
            Construct localized interaction graphs, aggregated traffic states,
            observations $o_{i,\kappa}^{(m)}$, and virtual proxies for
            lane-frontmost vehicles
            }

            \STATE Set $V_{i,T_m}^{(m)}(\cdot)\leftarrow0$

            \FOR{$\kappa=T_m-1$ down to $0$}

                \STATE \parbox[t]{0.82\linewidth}{
                Generate candidate trajectories $\Omega_{i,\kappa}^{(m)}$
                and reward weights $\boldsymbol{\theta}_{i,\kappa}^{(m)}$
                using Eq.~\eqref{eq:context_encoder}
                }

                \STATE \parbox[t]{0.82\linewidth}{
                Initialize $\pi_{i,\kappa}^{(0,m)}$ uniformly over
                the candidate actions
                }

                \FOR{$q=0$ to $N_{\mathrm{resp}}-1$}

                    \STATE \parbox[t]{0.82\linewidth}{
                    Evaluate $Q_{i,\kappa}^{(q,m)}$ using Eq.~\eqref{eq:explicit_soft_bellman},
                    with continuation values recursively computed at the
                    successor states induced by candidate joint actions
                    }

                    \STATE \parbox[t]{0.82\linewidth}{
                    Update $\pi_{i,\kappa}^{(q+1,m)}$ and
                    $V_{i,\kappa}^{(q+1,m)}$ using Eq.~\eqref{eq:explicit_logit_response}
                    and Eq.~\eqref{eq:explicit_soft_value}
                    }

                \ENDFOR

                \STATE \parbox[t]{0.82\linewidth}{
                Set $Q_{i,\kappa}^{(m)}$, $\pi_{i,\kappa}^{(m)}$,
                and $V_{i,\kappa}^{(m)}$ as the finalized response quantities
                }

                \STATE \parbox[t]{0.82\linewidth}{
                Match each valid demonstrated trajectory to the candidate
                with the closest terminal displacement to obtain
                $a_{i,\kappa}^{*(m)}$
                }

            \ENDFOR

            \STATE \parbox[t]{0.82\linewidth}{
            Compute $\mathcal{L}_{\mathrm{MAIRL}}^{(m)}$ using
            Eq.~\eqref{eq:expert_loglikelihood}
            and Eq.~\eqref{eq:episode_mairl_loss}
            }

        \ENDFOR

        \STATE \parbox[t]{0.82\linewidth}{
        Compute $\mathcal{L}_{\mathrm{MAIRL}}$
        using Eq.~\eqref{eq:mairl_loss_final}
        }

        \STATE \parbox[t]{0.82\linewidth}{
        Construct feedback preference pairs
        $\mathcal{C}_{i,\mathrm{pref}}^{\mathcal{B}}$
        using Eq.~\eqref{eq:preference_pair_set}
        }

        \STATE \parbox[t]{0.82\linewidth}{
        Evaluate $U_i^\chi=r_i(s^\chi,\boldsymbol{a}^{*,\chi})$
        and compute $\mathcal{L}_{\mathrm{pref}}$ using
        Eqs.~\eqref{eq:pref_loss_agent}--\eqref{eq:pref_loss}
        }

        \STATE \parbox[t]{0.82\linewidth}{
        Compute $\mathcal{L}_{\mathrm{reg}}$ using Eq.~\eqref{eq:reg_loss}
        and joint loss $\mathcal{L}$ using Eq.~\eqref{eq:joint_objective}
        }

        \STATE \parbox[t]{0.82\linewidth}{
        Update $\boldsymbol{\eta}$ by back-propagation through
        the unrolled marginal-response inference
        }

    \ENDFOR
\ENDFOR

\end{algorithmic}
\end{algorithm}

\section{Results} \label{sec:result}

In this section, we present experiments evaluating the proposed HFG-MAIRL method.
Section~\ref{sec:benchmarks} introduces the benchmark algorithms.
Section~\ref{sec:implementation} describes the dataset and implementation details.
Section~\ref{sec:performance} evaluates HFG-MAIRL against representative IRL baselines and analyzes the learned reward structure.
Sections~\ref{sec:ablation} and~\ref{sec:sensitivity} present the ablation study and sensitivity analysis of key components and parameter settings, respectively.

\subsection{Benchmarks} \label{sec:benchmarks}

The following reward-learning benchmarks are compared to evaluate the proposed method.

\begin{itemize}

    \item \textbf{Adversarial IRL (AIRL)}~\cite{fu2017learning}: This benchmark formulates IRL as an adversarial learning problem, where a discriminator distinguishes expert trajectories from generated trajectories and induces the reward from its output. AIRL enables expressive reward learning without relying on an explicitly parameterized feature-weight structure.

    \item \textbf{Margin IRL}~\cite{bashir2021inverse}: This benchmark adopts a margin-based reward-learning objective that maximizes the difference between expert and candidate trajectory rewards using a squared hinge loss. Unlike the maximum-entropy principle, it directly enforces a reward margin between demonstrated and alternative driving decisions.

    \item \textbf{Variational Bayesian IRL}~\cite{wei2023bayesian}: This benchmark treats reward parameters as latent variables and approximates their posterior distribution through variational inference. It provides probabilistic reward recovery with uncertainty modeling but does not explicitly incorporate human feedback preference supervision or iterative neighboring-vehicle responses.

    \item \textbf{MaxEnt IRL}~\cite{huang2021driving}: This benchmark adopts a single-agent Maximum Entropy IRL formulation that treats surrounding vehicles as dynamic environmental factors without explicitly modeling their mutual decision responses. It represents a probabilistic reward-learning approach based solely on demonstrated driving behaviors.

    \item \textbf{HFG-MAIRL (Ours)}: The proposed method formulates interactive driving as a finite-horizon Markov game and learns context-dependent reward functions from multi-vehicle demonstrations. Directly interacting vehicles are modeled explicitly through marginal-response inference, while broader traffic influences are represented through aggregated traffic states and a passive virtual forward-safety proxy. Available human feedback is further incorporated as ordinal preference supervision for the learned reward.

\end{itemize}

The benchmark comparison evaluates whether explicitly modeling multi-vehicle responses and incorporating complementary human preference information improve reward-based reconstruction of demonstrated driving decisions.

\subsection{Experiment Setup and Implementation Details} \label{sec:implementation}

\begin{figure}[ht]
    \centering
    \begin{subfigure}{0.96\linewidth}
        \centering
        \includegraphics[width=\linewidth]{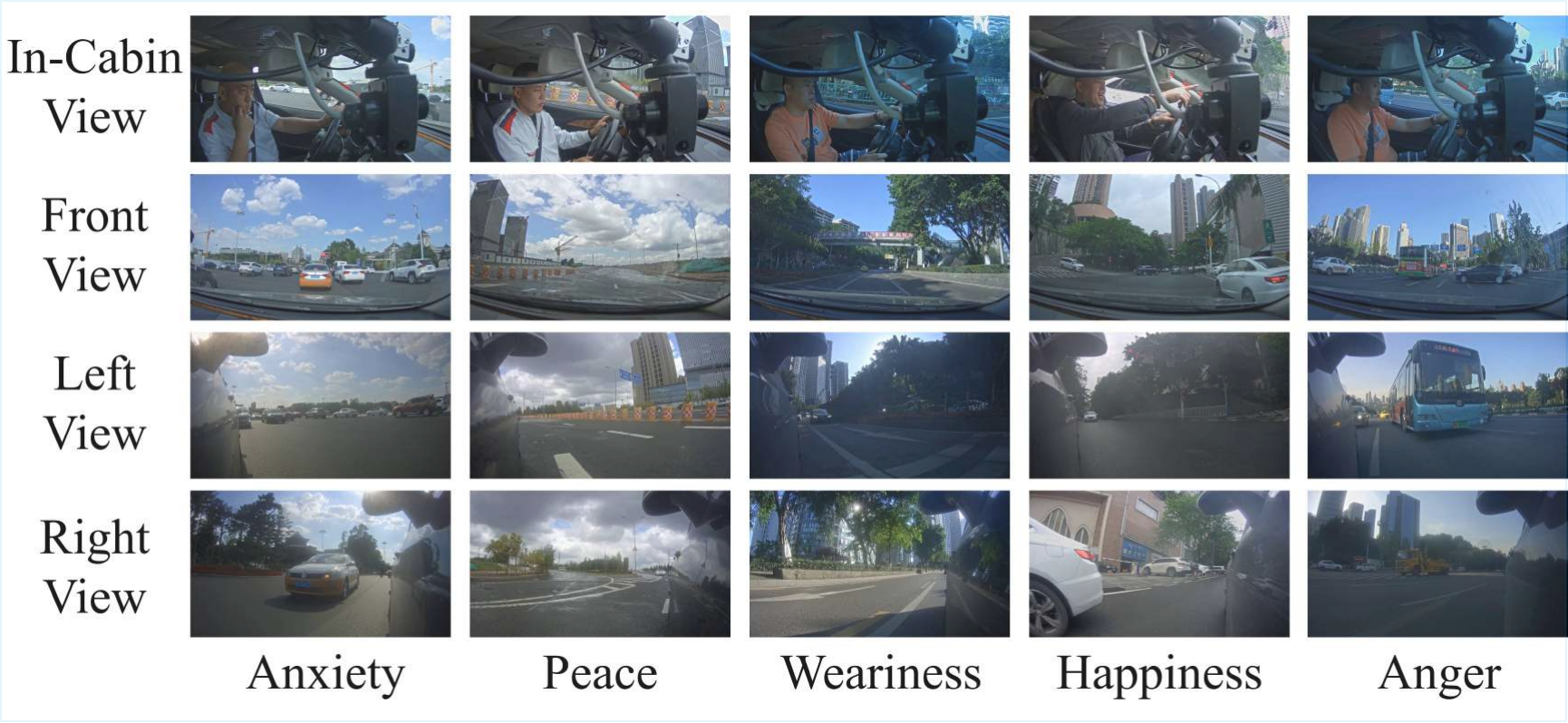}
        \label{fig:AIDE_Des}
    \end{subfigure}
    \caption{Illustration of the multimodal driver emotion dataset used in this work. AIDE captures naturalistic on-road driving with synchronized multi-view videos annotated with five emotion categories.}
    \label{fig:datasets_Des}
\end{figure}

In this study, we instantiate the general human-feedback formulation using driver emotion annotations. The proposed framework itself is not restricted to emotion feedback and can accommodate other categorical or ordinal feedback signals from which pairwise preferences can be constructed.

\subsubsection{Dataset}

We utilize the AIDE dataset~\cite{yang2023aide} to evaluate the proposed framework in naturalistic on-road driving scenarios.
Naturalistic datasets containing both traffic-scene observations and synchronized driver-state annotations are scarce, and AIDE provides paired observations of driver states and surrounding traffic through synchronized in-cabin, front, left, and right camera recordings.
Each sample consists of a 3-second driving clip with 45 frames per view and is annotated with an emotion category $e\in\{\textit{Anxiety},\textit{Peace},\textit{Weariness},\textit{Happiness},\textit{Anger}\}$.
Vehicle trajectories are obtained through YOLO-based detection and multi-object tracking~\cite{Jocher_Ultralytics_YOLO_2023} and projected into a bird's-eye-view coordinate system.
Only lane-changing and forward-driving scenes are included, as these scenarios contain explicit multi-vehicle interactions while providing sufficient trajectory observations for reward inference.

The dataset is split at the clip level into training, validation, and test sets with an approximate ratio of 70\%/15\%/15\%.
The validation set is used exclusively for model selection and configuration determination. After all model configurations and default hyperparameters are fixed, the test set is used for benchmark comparison, ablation studies, and sensitivity analysis.
Fig.~\ref{fig:datasets_Des} presents representative examples from the AIDE dataset.

\subsubsection{Implementation Details}

We implement all benchmark methods using the hyperparameter settings reported in their original publications, with minor adaptations to ensure compatibility with the action representation used in our evaluation. For a fair comparison, all methods use the same trajectory candidate generation procedure, terminal-anchor sets, observation horizons, and traffic observations. 

HFG-MAIRL uses an observation horizon and decision interval of $T_{\mathrm{obs}}=T_{\mathrm{dec}}=1.0\,\mathrm{s}$.
Candidate terminal anchors are uniformly discretized over longitudinal and lateral displacement ranges $[s_{\min},s_{\max}]$ and $[d_{\min},d_{\max}]$, whose bounds are estimated from the 5th and 95th percentiles of the training trajectories.
The same anchor specification is used across vehicles, and the default action space contains $K=20$ trajectory candidates.
Each candidate action corresponds to a 16-point trajectory segment over the subsequent decision interval.

For reward-feature construction, the reference speed is set to $v_{\mathrm{ref}}=15.0\,\mathrm{m/s}$.
The acceleration and yaw-rate normalization constants are $a_{\max}=3.0\,\mathrm{m/s^2}$ and $\omega_{\max}=0.5\,\mathrm{rad/s}$, respectively.
The anisotropic social-distance scales are set to $\sigma_x=10.0\,\mathrm{m}$ and $\sigma_y=2.0\,\mathrm{m}$, where $\sigma_y<\sigma_x$ increases sensitivity to lateral proximity during adjacent-lane interactions.

The context encoder $f_{\mathrm{ctx}}$ is implemented as a two-layer multilayer perceptron with a hidden dimension of 32 and outputs context-dependent weights for the individual and interaction reward features.
Multi-vehicle decision coupling is approximated using $N_{\mathrm{resp}}=3$ marginal-response iterations at each decision epoch.
Each episode contains $T_m=2$ decision intervals, with the discount factor and inverse-temperature parameter set to $\gamma=1.0$ and $\beta=1.0$, respectively.

The emotion scoring function $S(e)$ maps \textit{Happiness} to 2, \textit{Peace} to 1, and \textit{Anxiety}, \textit{Weariness}, and \textit{Anger} to 0.
These scores are used only to construct ordinal preference relations $\text{Happiness} \succ \text{Peace} \succ \{\text{Anxiety},~\text{Weariness}, \text{Anger}\}$ and are not treated as numerical reward targets. Any alternative numerical values that preserve this ordering would yield exactly the same set of preference pairs. Although the general formulation permits feedback for any subset $\mathcal{V}_{\mathrm{fb}}\subseteq\mathcal{V}$, AIDE provides emotion annotations only for the ego driver.
We therefore set $\mathcal{V}_{\mathrm{fb}}=\{\mathrm{ego}\}$ and apply the preference loss only to the reward of the feedback-annotated ego vehicle.

All learnable models are trained using the Adam optimizer with a learning rate of $0.001$ and a batch size of 8.
The default preference-loss and regularization weights are set to $\lambda_{\mathrm{pref}}=1.0$ and $\lambda_{\mathrm{reg}}=10^{-4}$, respectively.
All benchmark, ablation, and sensitivity experiments follow the same training and evaluation protocol.
All models are implemented in PyTorch and trained on a high-performance computing platform equipped with NVIDIA H100 and A40 GPUs.

\subsection{Performance Analysis} \label{sec:performance}

We evaluate HFG-MAIRL on the test set by measuring how accurately the learned policy reconstructs demonstrated vehicle behaviors. 
At each decision interval, the policy selects the highest-probability trajectory candidate for each vehicle based on vehicle states and traffic history updated using the observed trajectories.
The selected trajectories are compared with expert trajectories for all vehicles.
We report Average Displacement Error (ADE) and Final Displacement Error (FDE) for spatial accuracy, Speed Deviation Error (SDE) and Acceleration Deviation Error (AccDE) for kinematic accuracy, and Heading Deviation Error (HDE) for heading consistency. Results are reported as mean $\pm$ standard deviation over five random seeds.

Table~\ref{tab:quantitative_res} compares HFG-MAIRL with representative IRL benchmarks. HFG-MAIRL achieves the lowest reconstruction errors across all five metrics, with an ADE of 0.32 m, FDE of 0.95 m, SDE of 1.06 m/s, AccDE of 1.99 m/s$^2$, and HDE of 0.10 rad. Compared with AIRL, HFG-MAIRL reduces ADE, FDE, SDE, AccDE, and HDE by 85.5\%, 83.2\%, 68.3\%, 56.6\%, and 44.4\%, respectively. The other three IRL baselines achieve substantially lower errors than AIRL but remain considerably less accurate than HFG-MAIRL. In particular, compared with MaxEnt IRL, which achieves the strongest overall reconstruction performance among these baselines, HFG-MAIRL further reduces ADE by 76.5\%, FDE by 68.0\%, SDE by 36.1\%, AccDE by 24.3\%, and HDE by 28.6\%. 
The comparatively small performance differences among Margin IRL, Variational Bayesian IRL, and MaxEnt IRL, together with their larger gap from HFG-MAIRL, suggest that changing the conventional IRL learning objective alone is insufficient to achieve the reconstruction accuracy obtained by the proposed interaction-aware framework. 
The largest relative improvements of HFG-MAIRL are observed in ADE and FDE, indicating that its advantage is particularly pronounced in recovering the spatial outcome of demonstrated decisions. The simultaneous reductions in SDE, AccDE, and HDE further show that the improvement extends to the associated motion dynamics. 

\begin{table*}[htbp]
\centering
\caption{Comparison of trajectory reconstruction errors between HFG-MAIRL and representative IRL benchmarks.}
\label{tab:quantitative_res}
\renewcommand{\arraystretch}{1.25}
\setlength{\tabcolsep}{3pt}

\begin{tabular}{
>{\raggedright\arraybackslash}p{1.40in}
>{\centering\arraybackslash}p{0.86in}
>{\centering\arraybackslash}p{0.86in}
>{\centering\arraybackslash}p{0.90in}
>{\centering\arraybackslash}p{1.02in}
>{\centering\arraybackslash}p{0.88in}
}
\toprule
{\footnotesize\textbf{Benchmark}}
& {\footnotesize\textbf{ADE (m)}}
& {\footnotesize\textbf{FDE (m)}}
& {\footnotesize\textbf{SDE (m/s)}}
& {\footnotesize\textbf{AccDE (m/s$^2$)}}
& {\footnotesize\textbf{HDE (rad)}} \\
\midrule

AIRL
& $2.20 \pm 0.13$
& $5.65 \pm 0.35$
& $3.34 \pm 0.19$
& $4.58 \pm 0.19$
& $0.18 \pm 0.01$ \\

Margin IRL
& $1.43 \pm 0.17$
& $3.19 \pm 0.40$
& $1.75 \pm 0.22$
& $2.63 \pm 0.24$
& $0.16 \pm 0.01$ \\

Variational Bayesian IRL
& $1.39 \pm 0.17$
& $3.14 \pm 0.41$
& $1.75 \pm 0.22$
& $2.63 \pm 0.24$
& $0.14 \pm 0.01$ \\

MaxEnt IRL
& $1.36 \pm 0.17$
& $2.97 \pm 0.40$
& $1.66 \pm 0.21$
& $2.63 \pm 0.24$
& $0.14 \pm 0.01$ \\

\textbf{HFG-MAIRL}
& $\mathbf{0.32 \pm 0.03}$
& $\mathbf{0.95 \pm 0.11}$
& $\mathbf{1.06 \pm 0.11}$
& $\mathbf{1.99 \pm 0.19}$
& $\mathbf{0.10 \pm 0.01}$ \\
\bottomrule
\end{tabular}
\end{table*}

\begin{figure}[!t]
    \centering
    \includegraphics[width=1.00\linewidth]{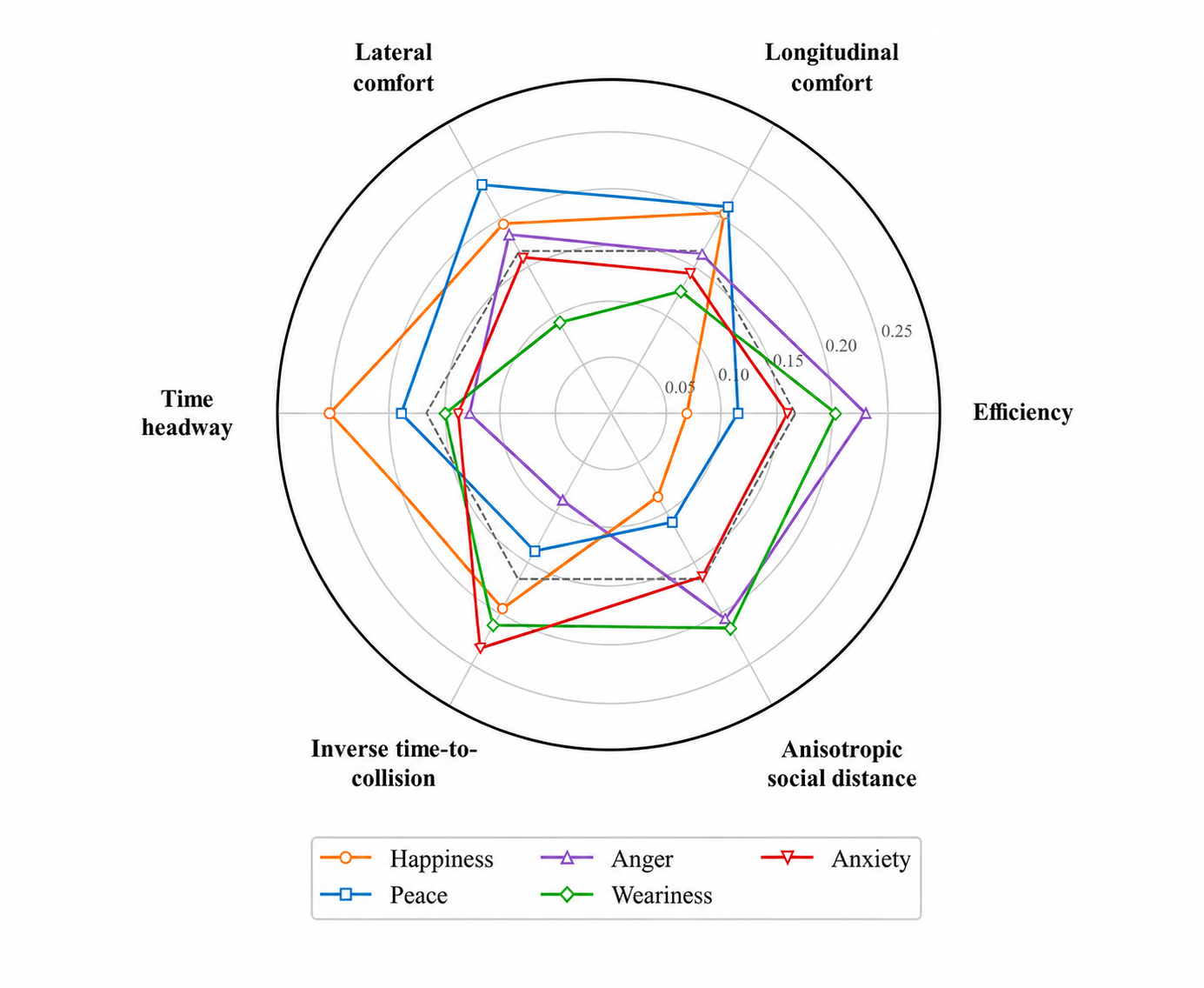}
    \caption{Relative composition of learned context-dependent reward weights for normalized features, grouped by emotion label.}
    \label{fig:emotion_relative_sensitivity_composition}
\end{figure}

\begin{figure}[!t]
    \centering
    \includegraphics[width=0.98\linewidth]{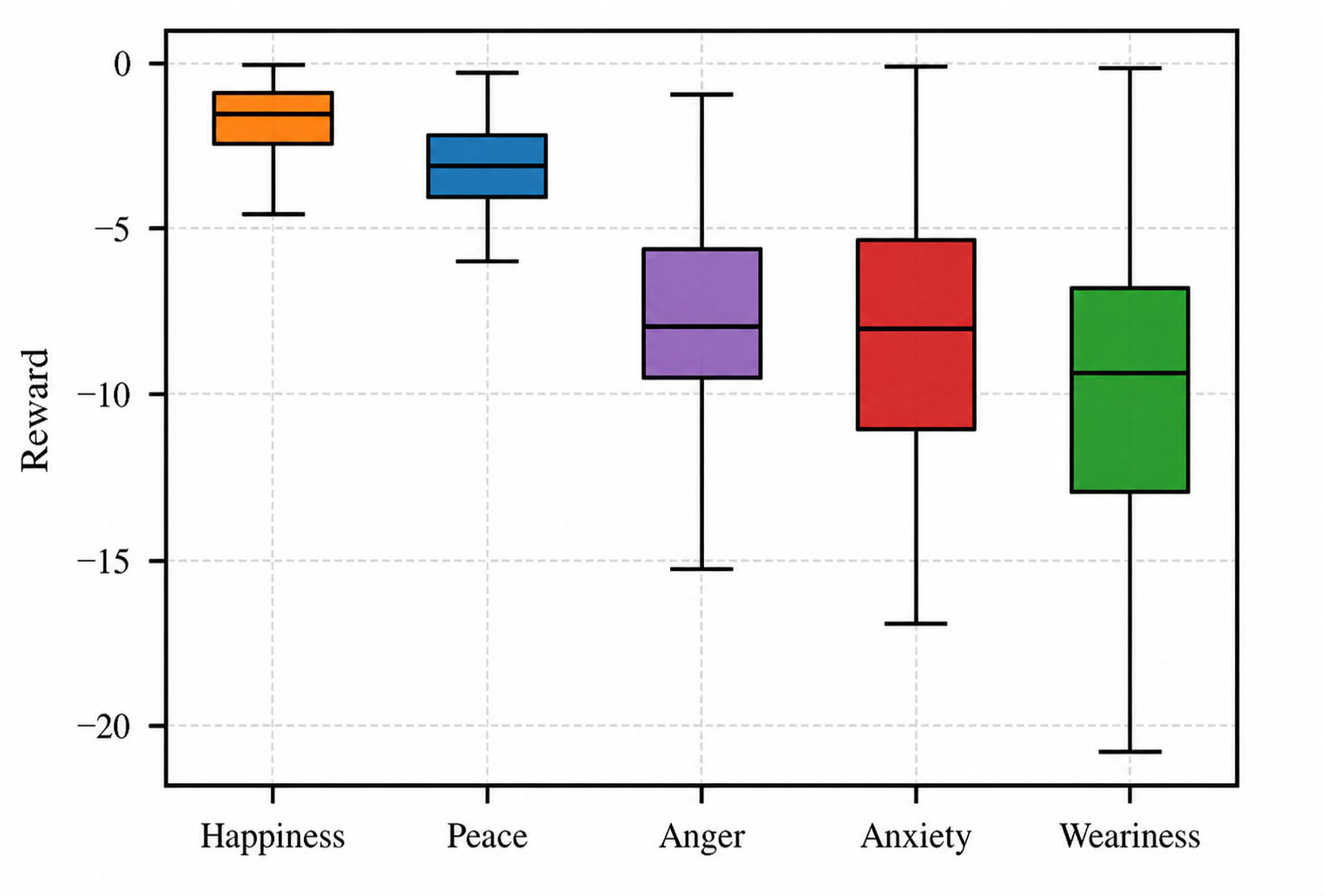}
    \caption{Learned reward distributions grouped by emotion label.}
    \label{fig:reward_monotonicity}
\end{figure}

\vspace{0.3em}
\noindent \emph{Value of human-feedback-guided reward interpretation.}
Beyond reconstructing demonstrated decisions, HFG-MAIRL aims to recover reward structures that reflect the interaction preferences associated with human feedback.
Figs.~\ref{fig:emotion_relative_sensitivity_composition} and~\ref{fig:reward_monotonicity} examine the learned reward structure for the feedback-annotated vehicle from two complementary perspectives.

Fig.~\ref{fig:emotion_relative_sensitivity_composition} compares the relative composition of the learned context-dependent reward-feature weights across emotion-labeled interaction contexts using the normalized features.
\emph{Happiness} places relatively greater emphasis on time headway, longitudinal comfort, and inverse time-to-collision, whereas \emph{Peace} is characterized primarily by lateral and longitudinal comfort.
\emph{Anxiety} assigns a relatively large weight to inverse time-to-collision, indicating greater relative emphasis on imminent closing risk.
\emph{Weariness} places greater emphasis on anisotropic social distance and inverse time-to-collision, while \emph{Anger} assigns relatively high weights to efficiency and anisotropic social distance.
These differences show that the context-dependent reward model does not associate all driving situations with a fixed preference structure, but instead captures diverse trade-offs across interaction contexts associated with different human responses.

Fig.~\ref{fig:reward_monotonicity} provides a complementary view by grouping the learned interval-level rewards of demonstrated vehicle decisions according to emotion label.
\emph{Happiness} and \emph{Peace} generally receive higher learned rewards, whereas \emph{Anger}, \emph{Anxiety}, and \emph{Weariness} exhibit lower and more dispersed reward distributions.
This overall separation is consistent with the ordinal preference construction, where \emph{Happiness} and \emph{Peace} receive higher feedback scores than the three lower-score emotion categories.
The three lower-score categories are not required to exhibit a strict internal ordering because they share the same preference score and therefore do not form preference pairs with one another.
Although \emph{Anger}, \emph{Anxiety}, and \emph{Weariness} share the same preference score and receive no explicit ordering constraints, their reward distributions still differ, indicating that the behavior-based MAIRL objective and context-dependent reward model capture additional preference structure beyond the ordering explicitly imposed by human feedback.

\subsection{Ablation Study}
\label{sec:ablation}

We evaluate the contributions of four key components in HFG-MAIRL through ablation experiments: the aggregated traffic state, virtual forward-safety proxy, marginal-response inference, and human feedback supervision.
Each component is removed individually, with the remaining model structure and evaluation settings kept unchanged. Table~\ref{tab:component_evidence} summarizes the resulting performance.

\vspace{0.3em}
\noindent \emph{Value of the aggregated traffic state.}
We evaluate the contribution of the aggregated traffic state to vehicle observations by removing it from the context representation while retaining its use for constructing the virtual forward-safety proxy. Without this component, ADE increases from 0.32 m to 0.34 m, FDE from 0.95 m to 0.97 m, SDE from 1.06 m/s to 1.09 m/s, AccDE from 1.99 m/s$^2$ to 2.03 m/s$^2$, and HDE from 0.10 rad to 0.11 rad. The degradation is modest but consistent across all five metrics. This pattern is consistent with the role of the aggregated traffic state as a complementary representation of broader traffic conditions: directly interacting neighbors already provide detailed vehicle-level interaction information, while the aggregated state supplies additional context from traffic beyond the explicitly modeled local interactions. Its removal therefore does not eliminate the primary interaction information, but reduces the broader contextual information available for context-dependent reward inference.

\vspace{0.3em}
\noindent \emph{Value of the virtual forward-safety proxy.}
We evaluate the contribution of the virtual proxy by removing it from HFG-MAIRL. The virtual proxy provides a forward-safety reference for a lane-frontmost vehicle when no physical leader is explicitly represented in the localized interaction graph. Removing this component increases ADE from 0.32 m to 0.33 m, SDE from 1.06 m/s to 1.09 m/s, AccDE from 1.99 m/s$^2$ to 2.03 m/s$^2$, and HDE from 0.10 rad to 0.11 rad, while FDE remains unchanged at 0.95 m at the reported precision. The relatively small overall effect is consistent with the specialized role of the virtual proxy: it is activated only when an explicitly represented physical leader is unavailable and contributes only to the forward-safety components of the interaction reward. The modest degradation therefore suggests that the proxy supplements the localized interaction model with useful forward-traffic constraints without becoming a dominant source of decision information.

\vspace{0.3em}
\noindent \emph{Value of marginal-response inference.}
We evaluate the contribution of marginal-response inference by disabling the iterative updates of neighboring-vehicle responses while retaining the remaining interaction components. Removing marginal-response inference increases ADE from 0.32 m to 0.47 m, FDE from 0.95 m to 1.40 m, SDE from 1.06 m/s to 1.61 m/s, AccDE from 1.99 m/s$^2$ to 2.90 m/s$^2$, and HDE from 0.10 rad to 0.15 rad. The substantial and consistent degradation across spatial, kinematic, and heading metrics indicates that iterative response modeling affects the overall evaluation of decisions. In HFG-MAIRL, each vehicle evaluates its candidate actions under the current response distributions of the other interacting vehicles, and the resulting policies are iteratively updated. Disabling this process removes the propagation of mutual decision responses across vehicles, leading to less accurate reconstruction of the demonstrated interactive decisions. These results provide empirical support for explicitly accounting for coupled multi-vehicle responses during reward inference.

\vspace{0.3em}
\noindent \emph{Value of human feedback preference supervision.}
We evaluate the contribution of human feedback supervision by removing the preference loss from HFG-MAIRL. Without human feedback supervision, ADE increases from 0.32 m to 0.46 m, FDE from 0.95 m to 1.36 m, SDE from 1.06 m/s to 1.56 m/s, AccDE from 1.99 m/s$^2$ to 2.81 m/s$^2$, and HDE from 0.10 rad to 0.13 rad. Notably, the human feedback does not directly supervise trajectory positions or kinematic quantities, and it provides only ordinal constraints on the rewards associated with demonstrated decisions. The broad degradation observed after removing this supervision indicates that these additional reward-ordering constraints help distinguish among behaviorally plausible reward structures and improve the resulting evaluation of trajectory candidates. Together with the emotion-grouped reward structures shown in Figs.~\ref{fig:emotion_relative_sensitivity_composition} and~\ref{fig:reward_monotonicity}, the results show that human feedback is a complementary source of supervision that contributes both to reward interpretation and to behavioral reconstruction beyond trajectory demonstrations alone.

\begin{table*}[htbp]
\centering
\caption{Comparison of trajectory reconstruction errors for HFG-MAIRL with different component ablations.}
\label{tab:component_evidence}
\renewcommand{\arraystretch}{1.25}
\setlength{\tabcolsep}{2.5pt}

\begin{tabular}{
>{\raggedright\arraybackslash}p{1.55in}
>{\centering\arraybackslash}p{0.86in}
>{\centering\arraybackslash}p{0.86in}
>{\centering\arraybackslash}p{0.90in}
>{\centering\arraybackslash}p{1.02in}
>{\centering\arraybackslash}p{0.88in}
}
\toprule
{\footnotesize\textbf{Configuration}}
& {\footnotesize\textbf{ADE (m)}}
& {\footnotesize\textbf{FDE (m)}}
& {\footnotesize\textbf{SDE (m/s)}}
& {\footnotesize\textbf{AccDE (m/s$^2$)}}
& {\footnotesize\textbf{HDE (rad)}} \\
\midrule

w/o Aggregated Traffic State
& $0.34 \pm 0.04$
& $0.97 \pm 0.10$
& $1.09 \pm 0.11$
& $2.03 \pm 0.17$
& $0.11 \pm 0.01$ \\

w/o Virtual Proxy
& $0.33 \pm 0.03$
& $0.95 \pm 0.11$
& $1.09 \pm 0.11$
& $2.03 \pm 0.19$
& $0.11 \pm 0.01$ \\

w/o Marginal Response
& $0.47 \pm 0.05$
& $1.40 \pm 0.17$
& $1.61 \pm 0.17$
& $2.90 \pm 0.29$
& $0.15 \pm 0.01$ \\

w/o Human Feedback
& $0.46 \pm 0.04$
& $1.36 \pm 0.13$
& $1.56 \pm 0.15$
& $2.81 \pm 0.25$
& $0.13 \pm 0.01$ \\

\textbf{HFG-MAIRL}
& $\mathbf{0.32 \pm 0.03}$
& $\mathbf{0.95 \pm 0.11}$
& $\mathbf{1.06 \pm 0.11}$
& $\mathbf{1.99 \pm 0.19}$
& $\mathbf{0.10 \pm 0.01}$ \\
\bottomrule
\end{tabular}
\end{table*}

\subsection{Sensitivity Analysis}
\label{sec:sensitivity}

We further evaluate the effects of two key parameters in HFG-MAIRL through sensitivity analysis: the number of trajectory candidates $K$ and the preference-loss weight $\lambda_{\mathrm{pref}}$.
Each parameter is varied individually, with the remaining model structure and evaluation settings kept unchanged.
Table~\ref{tab:sensitivity} summarizes the resulting performance. $\dagger$ indicates the default setting. 

\begin{table*}[htbp]
\centering
\caption{Sensitivity analysis of trajectory reconstruction errors for HFG-MAIRL with varying hyperparameters.}
\label{tab:sensitivity}
\renewcommand{\arraystretch}{1.25}
\setlength{\tabcolsep}{2.2pt}

\begin{tabular}{
>{\raggedright\arraybackslash}p{1.05in}
>{\centering\arraybackslash}p{0.55in}
>{\centering\arraybackslash}p{0.82in}
>{\centering\arraybackslash}p{0.82in}
>{\centering\arraybackslash}p{0.86in}
>{\centering\arraybackslash}p{0.98in}
>{\centering\arraybackslash}p{0.84in}
}
\toprule
{\footnotesize\textbf{Axis}}
& {\footnotesize\textbf{Setting}}
& {\footnotesize\textbf{ADE (m)}}
& {\footnotesize\textbf{FDE (m)}}
& {\footnotesize\textbf{SDE (m/s)}}
& {\footnotesize\textbf{AccDE (m/s$^2$)}}
& {\footnotesize\textbf{HDE (rad)}} \\
\midrule

$K$
& 10
& $0.42 \pm 0.02$
& $1.18 \pm 0.08$
& $1.40 \pm 0.09$
& $2.56 \pm 0.16$
& $0.13 \pm 0.01$ \\

$K$
& 20$\dagger$
& $0.32 \pm 0.03$
& $0.95 \pm 0.11$
& $1.06 \pm 0.11$
& $1.99 \pm 0.19$
& $\mathbf{0.10 \pm 0.01}$ \\

$K$
& 30
& $\mathbf{0.30 \pm 0.03}$
& $\mathbf{0.85 \pm 0.12}$
& $\mathbf{0.93 \pm 0.11}$
& $\mathbf{1.77 \pm 0.19}$
& $0.12 \pm 0.01$ \\

\midrule

$\lambda_{\mathrm{pref}}$
& 0.5
& $0.33 \pm 0.03$
& $0.96 \pm 0.11$
& $1.07 \pm 0.11$
& $2.01 \pm 0.19$
& $0.10 \pm 0.01$ \\

$\lambda_{\mathrm{pref}}$
& 1.0$\dagger$
& $\mathbf{0.32 \pm 0.03}$
& $\mathbf{0.95 \pm 0.11}$
& $\mathbf{1.06 \pm 0.11}$
& $\mathbf{1.99 \pm 0.19}$
& $\mathbf{0.10 \pm 0.01}$ \\

$\lambda_{\mathrm{pref}}$
& 2.0
& $0.33 \pm 0.03$
& $0.96 \pm 0.11$
& $1.09 \pm 0.11$
& $2.04 \pm 0.19$
& $0.10 \pm 0.01$ \\
\bottomrule
\end{tabular}
\end{table*}

\vspace{0.3em}
\noindent \emph{Sensitivity analysis on trajectory candidate number.}
We evaluate the effect of the number of trajectory candidates $K$ by varying the size of the discrete action space. Increasing $K$ from 10 to 20 improves all five reconstruction metrics, with ADE and FDE decreasing from 0.42 m and 1.18 m to 0.32 m and 0.95 m, respectively. This improvement is consistent with the action representation of HFG-MAIRL: a larger candidate set provides finer coverage of feasible terminal displacements and reduces the mismatch between demonstrated trajectories and the available discrete actions. Increasing $K$ further to 30 reduces ADE, FDE, SDE, and AccDE to 0.30 m, 0.85 m, 0.93 m/s, and 1.77 m/s$^2$, respectively, while HDE increases slightly from 0.10 rad to 0.12 rad. Thus, increasing candidate resolution generally improves spatial and kinematic reconstruction, although the benefit is not uniform across all motion characteristics. 
The default $K=20$ therefore provides strong reconstruction performance with a smaller action space, while $K=30$ offers additional accuracy on several metrics at the cost of evaluating more candidate actions during marginal-response inference.

\vspace{0.3em}
\noindent \emph{Sensitivity analysis on preference-loss weight.}
We evaluate the effect of the preference-loss weight $\lambda_{\mathrm{pref}}$ by varying the contribution of human-feedback-derived ordinal preference supervision during reward learning. Reconstruction performance remains relatively stable across $\lambda_{\mathrm{pref}}\in{0.5,1.0,2.0}$. The default $\lambda_{\mathrm{pref}}=1.0$ achieves the lowest ADE, FDE, SDE, and AccDE of 0.32 m, 0.95 m, 1.06 m/s, and 1.99 m/s$^2$, respectively, while HDE remains 0.10 rad across the three evaluated settings at the reported precision. The limited variation across this range indicates that the learned behavior is not highly sensitive to the exact weighting of the preference term. This stability is consistent with the joint learning objective, in which the MAIRL likelihood continues to ground the reward in demonstrated driving decisions while the preference term provides complementary ordinal supervision. 
The default $\lambda_{\mathrm{pref}}=1.0$ achieves the lowest or tied-lowest reconstruction errors across the evaluated test-set settings, while the limited variation across the three settings indicates robustness to moderate changes in the strength of human feedback supervision.

\section{Conclusion} 
\label{sec:conclusion}

In this paper, we propose HFG-MAIRL, a human-feedback-guided multi-agent inverse reinforcement learning framework for interactive driving behavior modeling. 
HFG-MAIRL formulates interactive driving as a finite-horizon Markov game and learns context-dependent reward functions from multi-vehicle demonstrations. It models coupled vehicle interactions through iterative marginal-response inference, represents broader traffic influences using aggregated traffic states, introduces a virtual forward-safety proxy when no physical leader is explicitly represented, and incorporates available human feedback as complementary ordinal preference supervision beyond trajectory demonstrations.
Experimental results on the AIDE dataset show that HFG-MAIRL consistently improves reconstruction accuracy over representative IRL baselines across spatial, kinematic, and heading metrics.
The learned reward-feature compositions and emotion-grouped reward distributions further reveal interpretable variations in driving preferences across interaction contexts associated with different human responses. 

This work suggests several directions for future research. First, we plan to integrate the learned human reward model into large-scale mixed-traffic simulations to investigate how human-aligned driving preferences influence traffic-level efficiency, safety, and interaction dynamics. Second, we plan to incorporate continuous multimodal human-state signals, such as physiological and behavioral responses, to provide more detailed preference information and better reflect how human responses change over time during complex driving interactions.

\bibliographystyle{IEEEtran}
\bibliography{ref.bib}

\begin{IEEEbiography}
[{\includegraphics[width=1in,height=1.25in,clip,keepaspectratio]{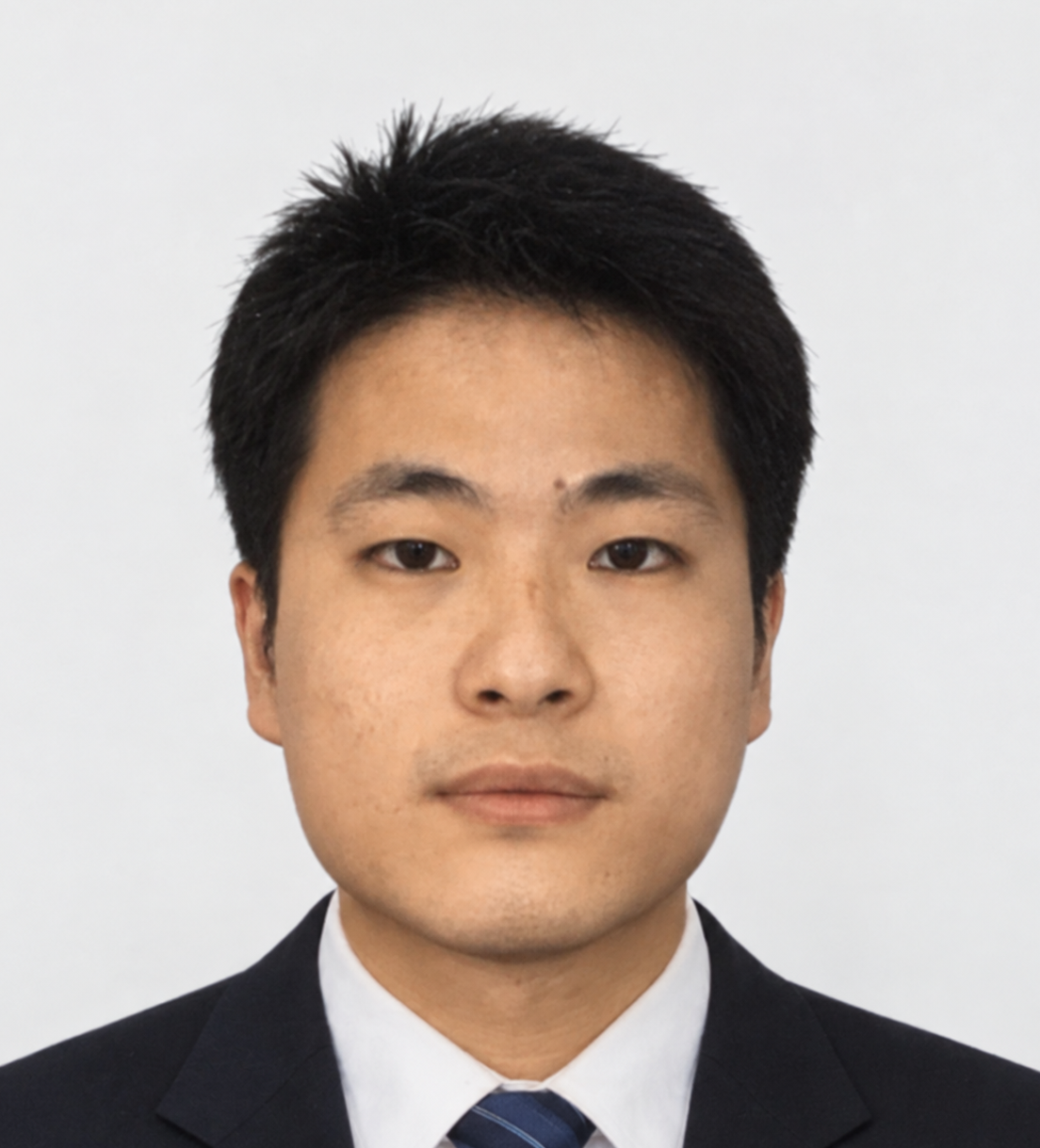}}]
{Longhao Yan} received the B.Eng. degree and M.Eng. degree in School of Electronics and Control Engineering from Chang’an University, Xi’an, China, in 2019 and 2022 respectively. He is currently working towards a Ph.D. degree with the National University of Singapore. His research interests include vehicle control and trajectory prediction of intelligent transportation system.
\end{IEEEbiography}
\begin{IEEEbiography}
[{\includegraphics[width=1in,height=1.25in,clip,keepaspectratio]{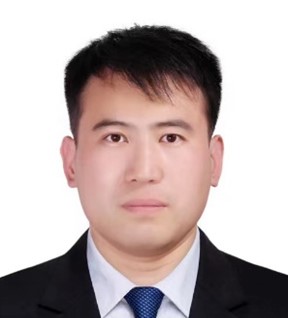}}]{Jinhao Liang} received the B.S. degree from School of Mechanical Engineering, Nanjing University of Science and Technology, Nanjing, China, in 2017, and Ph.D. degree from School of Mechanical Engineering, Southeast University, Nanjing, China, in 2022. Now he is a Research Fellow with Department of Civil and Environmental Engineering, National University of Singapore. His research interests include vehicle dynamics and control, and autonomous vehicles.
\end{IEEEbiography}
\begin{IEEEbiography}
[{\includegraphics[width=1in,height=1.25in,clip,keepaspectratio]{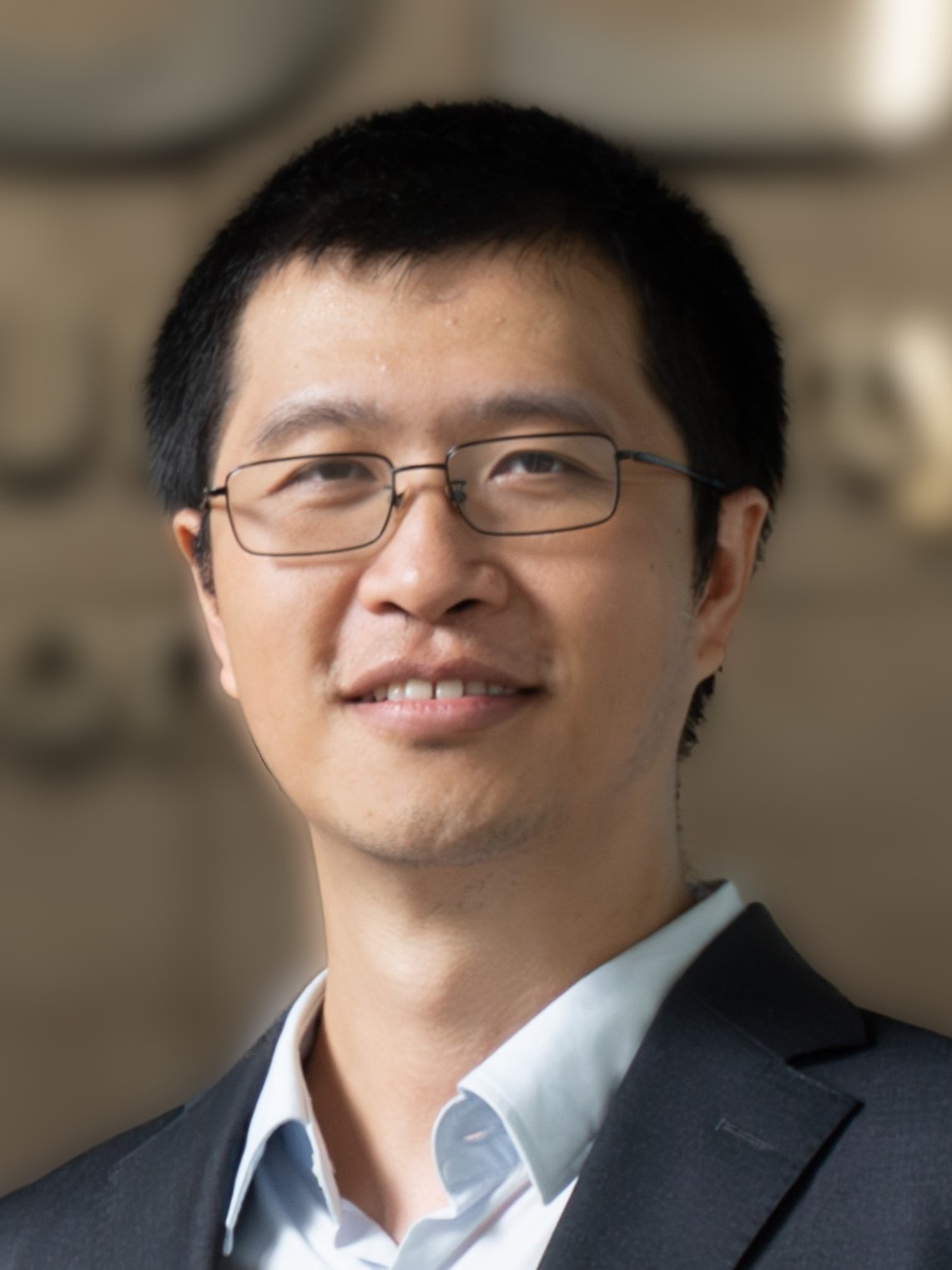}}]{Kaidi Yang} (Member, IEEE) is an Assistant Professor in the Department of Civil and Environmental Engineering at the National University of Singapore. Prior to this, he was a postdoctoral researcher with the Autonomous Systems Lab at Stanford University. He obtained a PhD degree from ETH Zurich and M.Sc. and B.Eng. degrees from Tsinghua University. His main research interest is the operation of future mobility systems enabled by connected and automated vehicles (CAVs) and shared mobility.
\end{IEEEbiography}

\end{document}